\documentclass[lettersize,journal]{IEEEtran}
\usepackage{amsmath,amsfonts}
\usepackage{algorithmic}
\usepackage{algorithm}
\usepackage{array}
\usepackage[caption=false,font=normalsize,labelfont=sf,textfont=sf]{subfig}
\usepackage{textcomp}
\usepackage{stfloats}
\usepackage{url}
\usepackage{verbatim}
\usepackage{graphicx}
\usepackage{cite}
\usepackage{amsmath}
\usepackage{amssymb}
\usepackage{booktabs}
\usepackage{multirow}
\usepackage{multicol}
\usepackage{colortbl}

\usepackage{arydshln}
\usepackage{xcolor}
\usepackage{tabularray}
\usepackage{hyperref}
\usepackage{color,soul}

\begin{document}
\title{Vision Also You Need: Navigating Out-of-Distribution Detection with
 Multimodal Large Language Model}

\author{ 
Haoran Xu$^{\ast}$, 
Yanlin Liu$^{\ast}$, 
Zizhao Tong$^{\ast}$, 
Jiaze Li,
Kexue Fu,
Yuyang Zhang,
\\Longxiang Gao,~\IEEEmembership{Senior Member,~IEEE,}
Shuaiguang Li,
Xingyu Li,
Yanran Xu,
Changwei Wang
\thanks{$^{\dag}$ Changwei Wang is the corresponding author (changweiwang@sdas.org).}
\thanks{$^{\ast}$Equal contribution.}
\thanks{
Haoran Xu is with Zhejiang University. Yanlin Liu is with Tsinghua University.
Zizhao Tong is with University of Chinese Academy of Sciences. Jiaze Li is with Zhejiang University.
Kexue Fu, Longxiang Gao, Shuaiguang Li and Changwei Wang are with the Key Laboratory of Computing Power Network and Information Security, Ministry of Education, Shandong Computer Science Center (National Supercomputer Center in Jinan), Qilu University of Technology (Shandong Academy of Sciences), Jinan, 250013, China; Shandong Provincial Key Laboratory of Computing Power Internet and Service Computing, Shandong Fundamental Research Center for Computer Science, Jinan, 250014, China; 
Yuyang Zhang is with University of Electronic Science and Technology of China.
Xingyu Li is with Mohamed bin Zayed University of Artificial Intelligence.
Yanran Xu is with RWTH Aachen University.
}
\thanks{This work was supported by the Shandong Provincial Key R\&D Program No.2025KJHZ013,  National Key R\&D Program of China No.2025YFE0216800, Shandong Provincial University Youth Innovation and Technology Support Program No.2022KJ291, Shandong Provincial Natural Science Foundation for Young Scholars Program No.ZR2025QC1627 and Qilu University of Technology (Shandong Academy of Sciences) Youth Outstanding Talent Program No. 2024QZJH02, Shandong Provincial Natural Science Foundation for Young Scholars (Category C) No.ZR2025QC1627.
}
}

\markboth{IEEE Transactions on Multimedia,~Vol.~XX, No.~XX, March~2026}%
{Shell \MakeLowercase{\textit{et al.}}: A Sample Article Using IEEEtran.cls for IEEE Journals}


\maketitle

\begin{abstract}
Out-of-Distribution (OOD) detection is a critical task that has garnered significant attention. The emergence of CLIP has spurred extensive research into zero-shot OOD detection, often employing a training-free approach. Current methods leverage expert knowledge from large language models (LLMs) to identify potential outliers. However, these approaches tend to over-rely on knowledge in the text space, neglecting the inherent challenges involved in detecting out-of-distribution samples in the image space. In this paper, we propose a novel pipeline, MM-OOD, which leverages the multimodal reasoning capabilities of MLLMs and their ability to conduct multi-round conversations for enhanced outlier detection. Our method is designed to improve performance in both near OOD and far OOD tasks. Specifically, (1) for near OOD tasks, we directly feed ID images and corresponding text prompts into MLLMs to identify potential outliers; and (2) for far OOD tasks, we introduce the sketch-generate-elaborate framework: first, we sketch outlier exposure using text prompts, then generate corresponding visual OOD samples, and finally elaborate by using multimodal prompts. Experiments demonstrate that our method achieves significant improvements on widely used multimodal datasets such as Food-101, while also validating its scalability on ImageNet-1K.
\end{abstract}


%
\IEEEpeerreviewmaketitle

\section{Introduction}
Amidst the continuous developments in deep learning, various  ~\cite{yang2019deep} expert models have achieved corresponding success across diverse domains. However, there is a rising focus on addressing model generalization in current research. It is well known that expert models in deep learning are often trained on closed datasets and then tested on expert-specific tasks which requires that training and testing datasets share identical distribution. Nevertheless, in complex open-world scenarios, the distribution of the training set and the test set is often considerable inconsistency, even substantial divergence, which often leads to the failure of the expert model in the real scenario. In some special scenarios, such as autonomous driving scenarios~\cite{ jin2022vwp}, the impact of these errors and failures can be profoundly serious. Therefore, it is equally important to identify out-of-distribution (OOD) samples and maintain good performance on ID samples.

Previous methods usually focus on detecting OOD examples only using closed dataset in training to predict OOD samples by lower confidence or higher energy. However, these methods are often less effective in the real scene. ATOM~\cite{chen2021atom} claimed that selecting more challenging outlier samples can help the model learn a better decision boundary~\cite{ ma2023feature} between ID and OOD. ID-like~\cite{bai2024idlike} proposed that the sampling ID-like samples come from the ID samples without auxiliary outliers dataset, because most challenging OOD samples can improve the performance. Recently, MCM~\cite{ming2022delving} introduces the zero-shot OOD detection method which borrows large-scale VLMs, i.e., CLIP~\cite{radford2021learning}. 
\begin{figure*}[!t]
    \centering
    \subfloat[ATOM]{
        \includegraphics[width=0.45\textwidth]{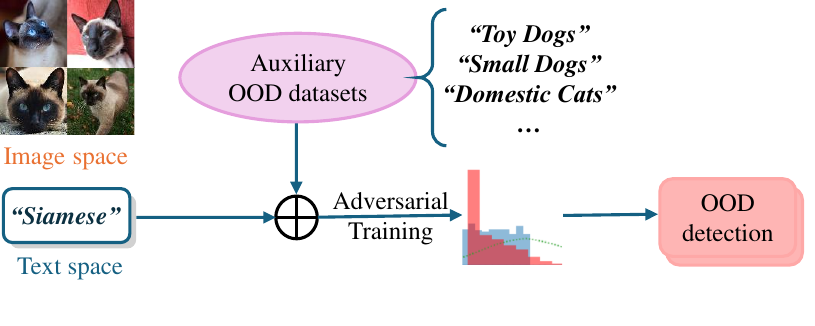}%
        \label{fig:atom}
    }
    \hfil
    \subfloat[ID-like]{
        \includegraphics[width=0.45\textwidth]{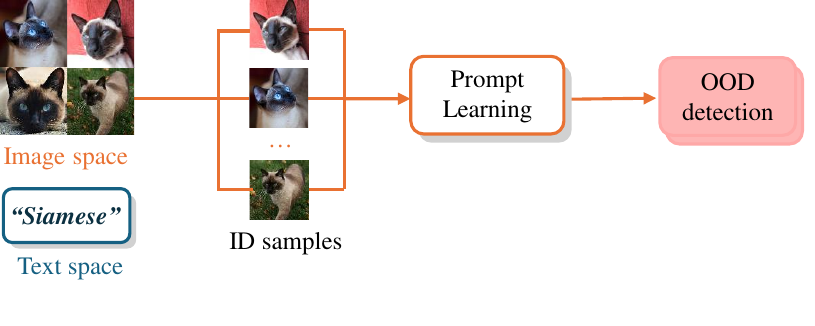}%
        \label{fig:idlike}
    }
    \\
    \subfloat[EOE]{
        \includegraphics[width=0.45\textwidth]{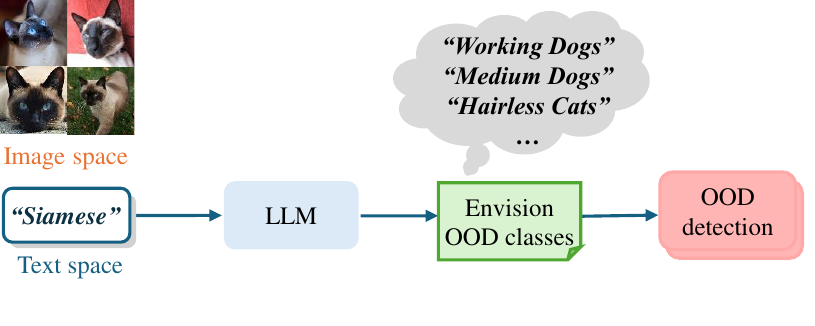}%
        \label{fig:eoe}
    }
    \hfil
    \subfloat[MM-OOD (Ours)]{
        \includegraphics[width=0.45\textwidth]{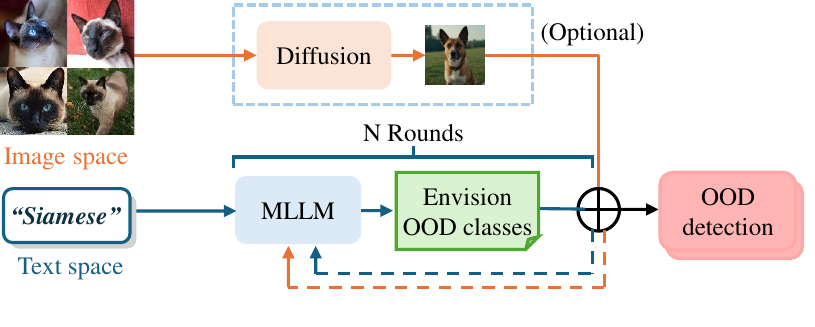}%
        \label{fig:mmood}
    }
    \caption{Illustrations of (a) ATOM, (b) ID-like, (c) EOE, and (d) MM-OOD (Ours). Previous methods such as (a), (b), and (c) typically utilized knowledge from a single modality (either text or vision) or relied on auxiliary datasets. Our approach stimulates the multimodal reasoning and multi-round dialogue capabilities of MLLMs.}
    \label{fig:methods_comparison}
\end{figure*}

Further, EOE~\cite{cao2024envisioning} extends the training-free framework ~\cite{min2017blind} by addressing the limitation of MCM in understanding out-of-distribution (OOD) samples. It leverages a large language model (LLM) to generate textual descriptions for OOD samples, thereby enhancing the model's ability to envision potential outlier exposure under a zero-shot OOD detection setting.

To sum up, the key to improve OOD detection is how to effectively generate outlier samples, regardless of whether the approach is training-free or involves fine-tuning. Hence, EOE attempts to leverage the reasoning ability of LLMs to address this challenge without other auxiliary datasets, but it overlooks one of the most remarkable properties of LLMs: multi-round conversations. Meanwhile, with the advancement of multimodal large language models (MLLMs), multimodal reasoning has demonstrated significant potential.

In this paper, we take a step further in EOE and employ multimodal large language models (MLLMs) to address this challenge. For near OOD task and far OOD task, we proposal a detailed and targeted pipeline which harnesses the Multimodal reasoning and Multi-round conversation capabilities ~\cite{zhu2023multi} to envision potential outlier exposure for Out-Of-Distribution detection, termed MM-OOD.

Technically, we straightforwardly feed ID images and corresponding text prompt into MLLMs for near OOD task to propose potential outlier exposure. Nonetheless, in far OOD tasks, we find that the outlier samples generated by MLLMs when feeding images tend to be very similar to the input image space, leading to the failure of the pipeline in handling far OOD tasks. Upon analysis, we infer that the outlier exposure provided by MLLMs is biased toward sampling near the input image space. When ID samples are fed into MLLMs, this bias restricts the model’s imaginative exploration of its stored knowledge. This observation motivates us to employ pretrained generative models to produce OOD samples into MLLMs. Thus, we design the sketch-generate-elaborate framework which is the form of multi-round conversations: sketch outlier exposure by text prompts, generate corresponding visual OOD samples, and elaborate by visual OOD samples and fine-grained text prompts.
Our contributions can be summarized as follows. First, we propose leveraging the multi-modal reasoning and multi-round conversation capability of multimodal large language models (MLLMs) to enhance outlier exposure. Second, we introduce MM-OOD, which incorporates differentiated designs for near OOD and far OOD tasks: for the former, we feed the in-distribution (ID) image and text prompt directly into MLLMs; for the latter, we propose a sketch-generate-elaborate framework. Third, our method MM-OOD consistently outperforms current state-of-the-art baselines across extensive experiments.

\section{Related Works}
\subsection{Traditional OOD Detection.}  
Traditional out-of-distribution (OOD) detection has been extensively studied in both visual~\cite{lin2021mood} and textual~\cite{ho2024longtailed, lang2024a, arora-etal-2021-types} classification tasks. Existing methods improve OOD detection through various strategies, including logit-based scoring~\cite{hendrycks17baseline, liu2020energy}, distance-based detectors~\cite{OECC2021}, robust representation learning~\cite{zhou-etal-2021-contrastive}, and pseudo-OOD sample generation~\cite{likehood2021}. These approaches can be broadly categorized into classification-based~\cite{Yuan_2024Discrimin, humblotrenaux2024noisy}, density-based~\cite{ren2019likelihood, xiao2020likelihood}, and reconstruction-based~\cite{zhou2022rethinking} techniques.

Classification-based methods rely on a well-trained in-distribution (ID) classifier and construct a scoring function to identify OOD samples. This scoring function may be derived from the input, hidden~\cite{sun2022out, lee2018simple, sun2022dice, zhu2023unleashing, wang2022watermarking}, output~\cite{Tang_2024_CVPR, hendrycks17baseline, liu2020energy}, or gradient spaces~\cite{huang2021importance, sharifi2024}. However, when the label space of the test data differs from that of the training data, conventional OOD detection methods typically require either retraining the model from scratch or fine-tuning, which incurs substantial computational overhead.

\subsection{Zero-Shot OOD Detection}
In the zero-shot setting, which leverages vision-language models (VLMs) without requiring further training on in-distribution (ID) data, zero-shot out-of-distribution OOD detection methods have shown promising results. However, recent works such as MCM~\cite{mcm2022} rely on closed-set ID labels and underutilize CLIP's open-world potential. ZOC~\cite{esmaeilpour2022zero} and CLIPN~\cite{2023clipn} address the open-world setting but require additional auxiliary datasets, limiting their scalability. Despite their effectiveness, these methods have not fully harnessed open-set OOD detection without additional auxiliary datasets. It constrains the future application and development of these methods in practice.

\begin{figure*}[!t]
    \centering
    \includegraphics[width=0.85\textwidth]{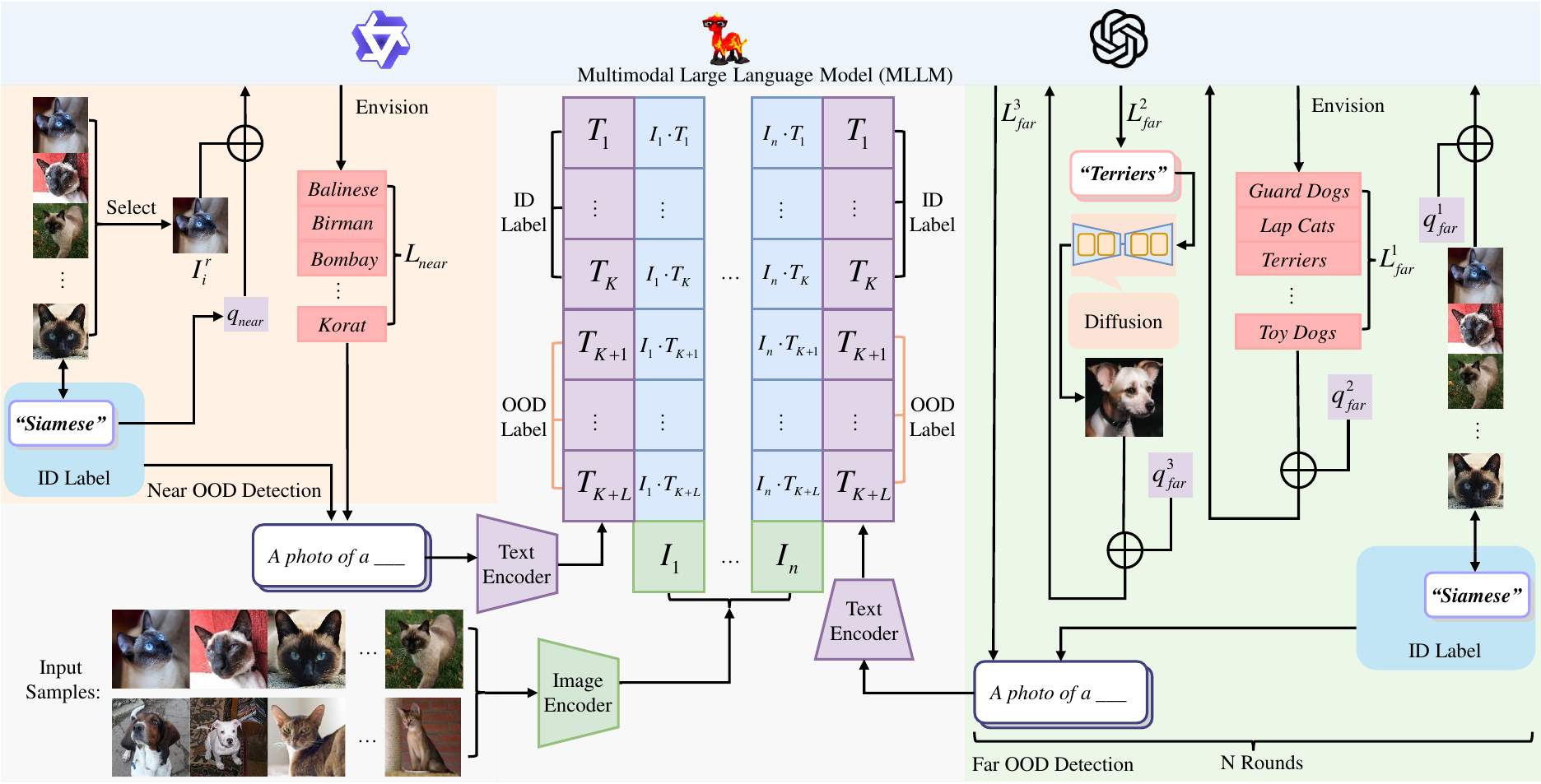}
    \caption{The pipeline of MM-OOD. We incorporate text prompts ($q_{\text{near}}$ for near-OOD detection and $q_{\text{far}}^{1}$--$q_{\text{far}}^{3}$ for far-OOD detection) and input images into MLLMs to generate potential out-of-distribution (OOD) class labels. These generated labels, combined with in-distribution (ID) class labels, are jointly encoded via a text encoder to construct a classifier for OOD detection.}
    \label{fig:pipeline}
\end{figure*}

\subsection{LLM-Guided Outlier Detection}
With the recent development of Large Language Models (LLMs), LLM-guided Outlier Detection methods are also emerging, such as EOE~\cite{cao2024envisioning}. EOE leveraged the expert knowledge and reasoning capabilities of LLMs to envision potential outliers and tackle the OOD detection problem. However, recent advancements in Multimodal Large Language Models (MLLMs)~\cite{li2022blip, alayrac2022flamingo} have led to improvements in multimodal reasoning~\cite{zhao2024lova3, zhao2024genixer}. EOE cannot fully utilize the potential of MLLM by using only textual reasoning capabilities, which limits the further improvement of the method. Our method, MM-OOD, innovatively uses MLLM-Guided Outlier Detection to address this limitation.

In summary, our approach MM-OOD focuses on zero-shot OOD detection task without auxiliary datasets with leveraging information from both textual and visual domains to facilitate the identification and analysis of potential outlier exposure. Figure.~\ref{fig:methods_comparison} shows the comparison between MM-OOD and other three representative methods.
\section{Method}
\label{sec:method}

\subsection{Preliminaries}
\noindent\textbf{Problem Setup.}
Out-of-distribution (OOD) detection can be formulated as a binary classification problem~\cite{Mandal_2019_CVPR}, where the goal is to determine whether a given input \( x \) originates from the in-distribution (ID) data distribution \( P_{\text{ID}} \) or an OOD distribution \( P_{\text{OOD}} \). This is mathematically expressed as:
\begin{equation}
D_{\theta}(x; P_{\text{ID}}, \mathcal{I}, \mathcal{T}) = \begin{cases} 
\text{ID} & S(x) \ge \theta \\
\text{OOD} & S(x) < \theta 
\end{cases}
\label{formula:dectector}
\end{equation}

where \( D(\cdot) \) is the OOD detector, \( x \) denotes the input image, \( x \in \mathcal{X} \), \( \mathcal{X} = \{\text{ID}, \text{OOD}\} \), and \( P_{\text{ID}} \) defines the space of ID class labels. The OOD detection score function \( S \) is derived from the similarity between the visual representation \( \mathcal{I}(x) \) and the textual representation \( \mathcal{T}(t) \), where \( t \) is the textual input to the text encoder, such as "a photo of a {ID class}", and \( \theta \) is the threshold to distinguish ID/OOD classes.

\noindent\textbf{OOD Detection Tasks.}
In this paper, we refer to EOE setting and categorize OOD detection tasks into two types: near and far OOD detection.
\begin{itemize}
    \item \textbf{Near OOD detection} focuses on identifying OOD classes that are closely related to ID classes, especially in visually similar categories, such as "dog" and "wolf".
    \item \textbf{Far OOD detection} This involves identifying OOD classes that are distant from ID classes within the label space and semantic space.
\end{itemize}

\subsection{MM-OOD}
In this section, we introduce our proposed method \textbf{MM-OOD}. The pipeline is shown in the Figure~\ref{fig:pipeline}. Overall, we divide MM-OOD into two steps: propose outlier class labels and calculate detection score based on outlier class labels. MM-OOD is mainly based on the first step to improve. Specifically, we design corresponding structures based on the characteristics of different tasks. 

For near OOD detection task, we directly feed ID images and corresponding text prompts into MLLMs to propose potential outlier exposure. On the other hand, for far OOD detection task, we propose the sketch-generate-elaborate framework to reason outlier classes. 

\noindent\textbf{Near OOD Detection Branch.}
For each ID label, it is necessary to select the most representative image. Firstly, we calculate the average feature vector by CLIP of all images. Here is corresponding equation:
\begin{equation}
\bar{f}_i = \frac{1}{N_i} \sum_{j=1}^{N_i} f(I_{ij}),
\end{equation}
where $N_i$ is the number of images for ID label $i$ and $f(I_{ij})$ is the feature vector from image $j$ of $i$th ID class label. 

Then, the image whose feature vector is closest to this average is then designated as the representative image for that label. 
\begin{equation}
I_{i}^r = \arg \min_{I_{ij} \in \mathcal{I}_i} \| f(I_{ij}) - \bar{f}_i \|,
\end{equation}
where $I_{i}^r$ is the representative  ID image in ID label. Subsequently, we straightforwardly feed the ID class label and the corresponding ID image into MLLM to generate $N_{o}$ outlier class labels. Assuming we have $K$ ID class labels, we will generate $N_{o} \times K = L$ outlier class labels. Details of the prompt input to MLLMs are presented in the Appendix.

\noindent\textbf{Far OOD Detection Branch.}
 For far OOD detection task, we conduct a series of experiments and explorations. We find that when we directly feed both the ID image and ID class label into the MLLM, following the pipeline for the near-OOD task, the results are below expectations. Through analysis, we believe that this is due to the induction bias of the MLLM. Compared to pure text input, feeding the ID image constrains the MLLM’s generative potential and causes it to sample near the ID image space, which hinders performance on far OOD detection tasks.

Based on the aforementioned points, in order to achieve far sampling of outlier samples, we first guide MLLM to summarize these classes into $M$ respective primary categories. All subsequent operations are conducted using these $M$ primary categories, allowing us to avoid being restricted to local optimal solutions.

\begin{figure}[t]
\centering
\includegraphics[width=0.85\linewidth]{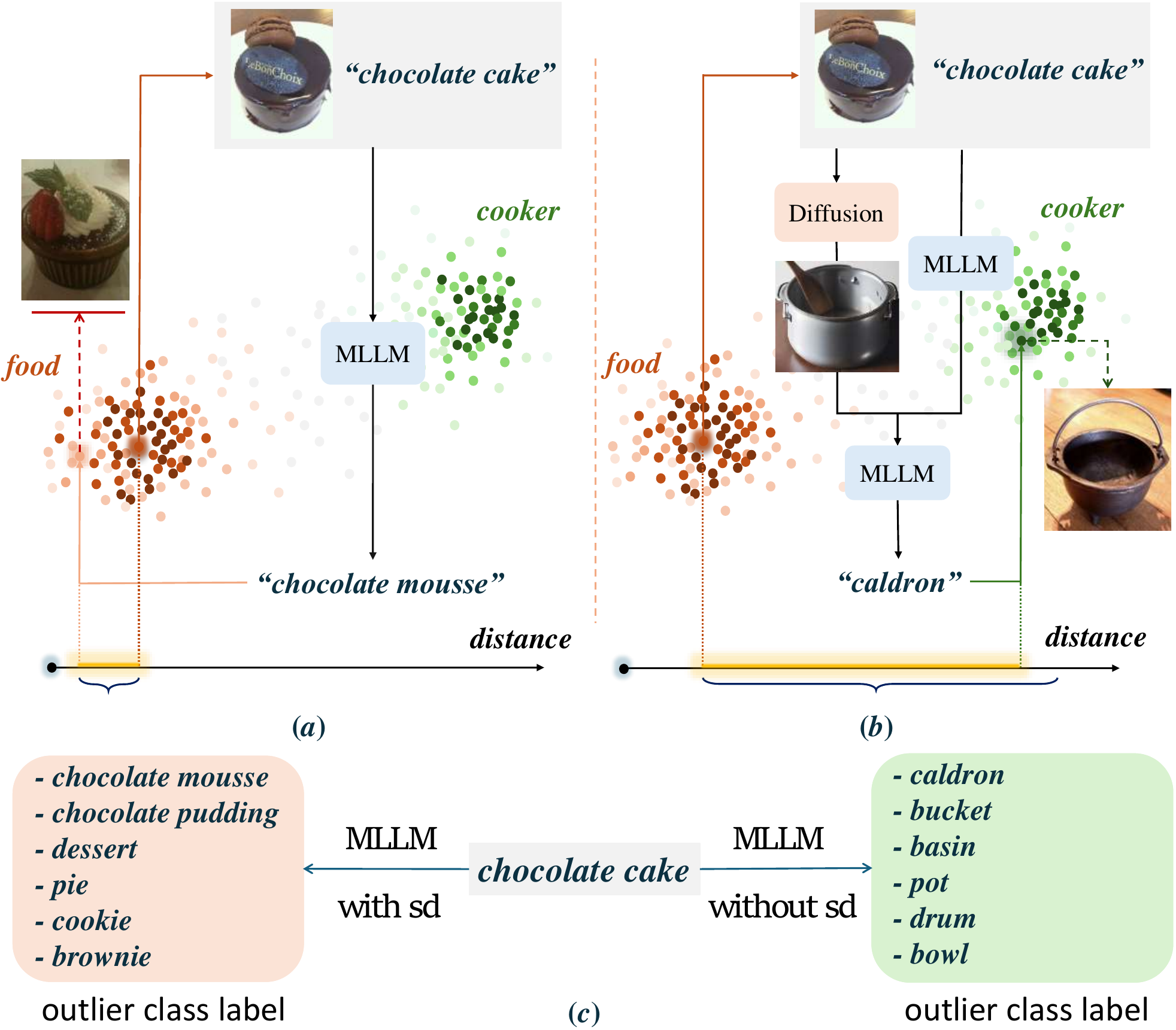}
\caption{Illustration of outlier class label generation: (a) without stable diffusion, (b) with stable diffusion, and (c) comparison between the two approaches.}
\label{fig:respren}
\end{figure}

Therefore, we consider generating out-of-distribution (OOD) images using generative models and feeding these OOD images into the MLLM instead of the ID images. The OOD images generated by the pretrained generative model serve as a demonstration of the OOD space in the image domain, guiding the MLLM to generate outlier class labels that are farther removed from the original ID space. Since the OOD images are distant from the ID image space, the MLLM shifts from near sampling to far sampling, which enhances MM-OOD's performance in far OOD detection tasks. Meanwhile, the OOD images input into the MLLM are maximally distinct from the ID image space, enabling our methodology to outperform our method with ID image input, EOE and random sampling method. Figure \ref{fig:respren} demonstrates this. Furthermore, we utilize the multi-round conversation capability of MLLM to propose the sketch-generate-elaborate framework. 

Specifically, the sketch operation initiates the MLLM to form a preliminary conception of the text label for the OOD sample. Subsequently, the generate operation directs the generative model to produce the corresponding OOD image based on this text label. Finally, the elaborate operation inputs the OOD image and the aforementioned text labels into the MLLM for the final generation. The framework essentially applies the multi-modal and multi-round dialogue characteristics of MLLM, while also conforming to the logic of the chain of thought (CoT)~\cite{wei2023chainofthoughtpromptingelicitsreasoning}. We give an in-depth analysis and a practical example in the section~\ref{in_depth_discussion}. The specific prompt is provided in Figure \ref{fig:converation}. The detailed algorithm flow is in the Appendix.

\begin{table*}[t]
\centering
\caption{Comparison results with MCM~\cite{ming2022delving}, MaxLogit~\cite{hendrycks2022scaling}, Energy~\cite{liu2020energy}, and the most recent EOE~\cite{cao2024envisioning} are reported under different datasets. \textbf{Bold} fonts highlight the best accuracy. $K$ denotes the number of classes in the corresponding in-distribution (ID) dataset.}
\label{tab:far-ood}
\footnotesize
\setlength{\tabcolsep}{4.2pt}
\renewcommand{\arraystretch}{1.05}
\begin{tabular}{lccccccccccccccc}
\toprule
        &  &\multicolumn{8}{c}{\textbf{OOD Dataset}} & \multicolumn{2}{c}{\multirow{2}{*}{\textbf{Average}}}  \\
        \multirow{2}{*}{\textbf{ID Dataset}} &\multirow{2}{*}{\textbf{Method}} & \multicolumn{2}{c}{iNaturalist} & \multicolumn{2}{c}{SUN} & \multicolumn{2}{c}{Places} & \multicolumn{2}{c}{Texture} & & &\\
          \cmidrule(r){3-4} \cmidrule(r){5-6} \cmidrule(r){7-8} \cmidrule(r){9-10} \cmidrule(r){11-12}  
		&  &\multicolumn{1}{c}{\textbf{FPR95}$\downarrow$} & \multicolumn{1}{c}{\textbf{AUROC}$\uparrow$} &
  \multicolumn{1}{c}{\textbf{FPR95}$\downarrow$} & 
  \multicolumn{1}{c}{\textbf{AUROC}$\uparrow$} & \multicolumn{1}{c}{\textbf{FPR95}$\downarrow$} &\multicolumn{1}{c}{\textbf{AUROC}$\uparrow$} & \multicolumn{1}{c}{\textbf{FPR95}$\downarrow$} & \multicolumn{1}{c}{\textbf{AUROC}$\uparrow$} & 
  \multicolumn{1}{c}{\textbf{FPR95}$\downarrow$} & \multicolumn{1}{c}{\textbf{AUROC}$\uparrow$}\\ 
		\cmidrule(r){1-1} \cmidrule(r){2-2}
  \cmidrule(r){3-4} \cmidrule(r){5-6} \cmidrule(r){7-8} \cmidrule(r){9-10} \cmidrule(r){11-12}  
\multirow{7}{*}{\textbf{CUB-200-2011}}
& MCM & 31.48 & 91.00 & 7.39 & 98.50 & 9.73 & 98.00 & 7.05 & 98.74 & 13.91 & 96.56 \\
& MaxLogit & 15.31 & 95.08 & 0.01 & 100.00 & 0.11 & 99.97 & 0.00 & 100.00 & 3.86 & 98.76 \\
& Energy & 15.92 & 94.92 & 0.01 & 100.00 & 0.10 & 99.97 & 0.02 & 100.00 & 4.01 & 98.72 \\
& EOE+LLaVA, ,L=4$\times K$ & 16.18 & 94.63 & 0.54 & 99.83 & 0.96 & 99.77 & 0.29 & 99.92 & 4.49 & 98.54 \\
& EOE+LLaVA, L=12$\times K$ & 14.80 & 95.02 & 0.16 & 99.95 & 0.30 & 99.92 & 0.02 & 99.99 & 3.82 & 98.72 \\
& Ours+LLaVA, ,L=4$\times K$ & 15.82 & 94.77 & 0.39 & 99.88 & 0.82 & 99.81 & 0.12 & 99.96 & 4.29 & 98.61 \\
& Ours+LLaVA, L=12$\times K$ & 15.23 & 94.86 & 0.17 & 99.95 & 0.33 & 99.92 & 0.02 & 99.99 & 3.94 & 98.68 \\
\midrule
\multirow{7}{*}{\textbf{Stanford Cars}}
& MCM & 1.22 & 99.52 & 0.13 & 99.92 & 0.56 & 99.81 & 0.02 & 99.97 & 0.48 & 99.81 \\
& MaxLogit & 0.00 & 100.00 & 0.60 & 99.85 & 0.90 & 99.74 & 0.00 & 100.00 & 0.38 & 99.90 \\
& Energy & 0.00 & 100.00 & 0.80 & 99.74 & 1.13 & 99.65 & 0.04 & 99.99 & 0.49 & 99.84 \\
& EOE+LLaVA, ,L=4$\times K$& 0.00 & 100.00 & 0.08 & 99.98 & 0.35 & 99.91 & 0.00 & 100.00 & 0.12 & 99.97 \\
& EOE+LLaVA, L=12$\times K$ & 0.00 & 100.00 & 0.11 & 99.98 & 0.36 & 99.91 & 0.00 & 100.00 & 0.12 & 99.97 \\
& Ours+LLaVA, ,L=4$\times K$ & 0.00 & 100.00 & 0.09 & 99.98 & 0.36 & 99.91 & 0.00 & 100.00 & 0.11 & 99.97 \\
& Ours+LLaVA, L=12$\times K$ & 0.00 & 100.00 & 0.12 & 99.97 & 0.36 & 99.91 & 0.00 & 100.00 & 0.12 & 99.97 \\
\midrule
\multirow{7}{*}{\textbf{Food-101}}
& MCM & 5.85 & 98.76 & 1.07 & 99.75 & 2.03 & 99.52 & 2.87 & 99.35 & 2.95 & 99.34 \\
& MaxLogit & 1.34 & 99.71 & 0.19 & 99.95 & 0.62 & 99.86 & 5.58 & 98.77 & 1.93 & 99.57 \\
& Energy & 1.31 & 99.70 & 0.22 & 99.94 & 0.61 & 99.86 & 8.59 & 98.07 & 2.68 & 99.39 \\
& EOE+LLaVA, ,L=4$\times K$ & 6.05 & 98.63 & 0.78 & 99.78 & 1.56 & 99.61 & 2.60 & 99.43 & 2.75 & 99.36 \\
& EOE+LLaVA, L=12$\times K$ & 4.62 & 99.00 & 0.66 & 99.82 & 1.33 & 99.67 & 2.27 & 99.50 & 2.22 & 99.50 \\
& Ours+LLaVA, ,L=4$\times K$ & 5.15 & 98.83 & 0.66 & 99.81 & 1.44 & 99.63 & 1.88 & 99.54 & 2.28 & 99.45 \\
& Ours+LLaVA, L=12$\times K$ & 2.00 & 99.60 & 0.28 & 99.92 & 0.86 & 99.79 & 1.34 & 99.69 & 1.12 & 99.75 \\
\midrule
\multirow{7}{*}{\textbf{Oxford Pets}}
& MCM & 19.48 & 96.43 & 1.17 & 99.73 & 2.20 & 99.52 & 1.18 & 99.72 & 6.01 & 98.85 \\
& MaxLogit & 0.04 & 99.97 & 0.07 & 99.98 & 0.30 & 99.92 & 0.39 & 99.89 & 0.20 & 99.94 \\
& Energy & 0.05 & 99.98 & 0.08 & 99.98 & 0.30 & 99.93 & 0.47 & 99.89 & 0.22 & 99.94 \\
& EOE+LLaVA, ,L=4$\times K$ & 0.36 & 99.85 & 0.73 & 99.81 & 1.29 & 99.66 & 0.48 & 99.86 & 0.71 & 99.79 \\
& EOE+LLaVA, L=12$\times K$ & 0.14 & 99.92 & 0.47 & 99.88 & 0.98 & 99.75 & 0.23 & 99.91 & 0.45 & 99.86 \\
& Ours+LLaVA, ,L=4$\times K$ & 0.11 & 99.94 & 0.43 & 99.90 & 0.83 & 99.79 & 0.21 & 99.93 & 0.40 & 99.89 \\
& Ours+LLaVA, L=12$\times K$ & 0.09 & 99.94 & 0.69 & 99.83 & 1.21 & 99.69 & 0.23 & 99.90 & 0.55 & 99.84 \\
\midrule
\multirow{7}{*}{\textbf{Average}}
& MCM & 14.51 & 96.43 & 2.44 & 99.48 & 3.63 & 99.21 & 2.78 & 99.45 & 5.84 & 98.64 \\
& MaxLogit & 4.17 & 98.69 & 0.22 & 99.95 & 0.48 & 99.87 & 1.49 & 99.67 & 1.59 & 99.54 \\
& Energy & 4.32 & 98.65 & 0.28 & 99.92 & 0.54 & 99.85 & 2.28 & 99.49 & 1.85 & 99.47 \\
& EOE+LLaVA, ,L=4$\times K$ & 5.65 & 98.28 & 0.53 & 99.85 & 1.04 & 99.74 & 0.84 & 99.80 & 2.02 & 99.42 \\
& EOE+LLaVA, L=12$\times K$ & 4.89 & 98.49 & 0.35 & 99.91 & 0.74 & 99.81 & 0.63 & 99.85 & 1.65 & 99.51 \\
& Ours+LLaVA, ,L=4$\times K$ & 5.27 & 98.39 & 0.39 & 99.89 & 0.86 & 99.79 & 0.55 & 99.86 & 1.77 & 99.48 \\
\rowcolor{gray!20}
& Ours+LLaVA, L=12$\times K$ & \textbf{4.33} & \textbf{98.60} & \textbf{0.32} & \textbf{99.92} & \textbf{0.69} & \textbf{99.83} & \textbf{0.40} & \textbf{99.90} & \textbf{1.43} & \textbf{99.56} \\
\bottomrule
\end{tabular}
\vspace{-0.2cm}
\end{table*}

\noindent\textbf{Detection Score Calculation.}
In this section, we enhance the separation between in-distribution (ID) and out-of-distribution (OOD) samples by computing a detection score. Building upon the outlier exposure strategy proposed in EOE~\cite{cao2024envisioning}, we leverage the generated outlier class labels to perform OOD detection. The detailed formulation is given below.

\begin{equation}
  s_ {i}  (x)=  \frac {\mathcal{I}(x)\cdot \mathcal{T}(t_ {i})}{|\mathcal{I}(x)|\cdot |\mathcal{T}(t_{i})|}, t_ {i}  \in \mathcal{Y}_{id} \cup  \mathcal{Y}_{outlier}
\end{equation}

\begin{equation}
\begin{aligned}
S(x) &=  \max_{i \in [1, K]} \frac{e^{s_i(x)}}{\sum_{j=1}^{K+L} e^{s_j(x)}} \\ 
&\quad - \beta \cdot \max_{k \in (K, K+L]} \frac{e^{s_k(x)}}{\sum_{j=1}^{K+L} e^{s_j(x)}}
\end{aligned}
\end{equation}
where $\beta$ is a hyperparameter. In our setting, we set it to 0.25. Finally, the OOD detector is followed by formula \ref{formula:dectector}.

\section{Experiments}
\subsection{Setups}
\noindent\textbf{Near OOD Detection.} We use ImageNet-10 and ImageNet-20 as ID and OOD datasets in a complementary manner, where ImageNet-10 serves as the ID dataset and ImageNet-20 as the OOD dataset, and vice versa. Both datasets are subsets of ImageNet-1K. ImageNet-10 is constructed to mimic the class distribution of CIFAR-10~\cite{krizhevsky2009learning}, while ImageNet-20 consists of 20 classes semantically similar to those in ImageNet-10. These datasets were introduced by MCM~\cite{ming2022delving}.

\noindent\textbf{Far OOD Detection.} For in-distribution (ID) datasets, we utilize Oxford-IIIT Pets~\cite{parkhi2012cats}, Stanford Cars~\cite{krause20133d}, CUB-200-2011~\cite{wah2011caltech}, Food-101~\cite{bossard14}, and ImageNet-1K~\cite{deng2009imagenet}. Among these, ImageNet-1K is used to demonstrate the scalability of our method on large-scale datasets. For out-of-distribution (OOD) evaluation, we employ large-scale datasets including Texture~\cite{cimpoi2014describing}, Places~\cite{zhou2017places}, iNaturalist~\cite{van2018inaturalist}, SUN~\cite{xiao2010sun}, and the OOD benchmark curated by MOS~\cite{huang2021mos}.

\begin{figure*}[t]
\centering
\includegraphics[width=0.8\linewidth]{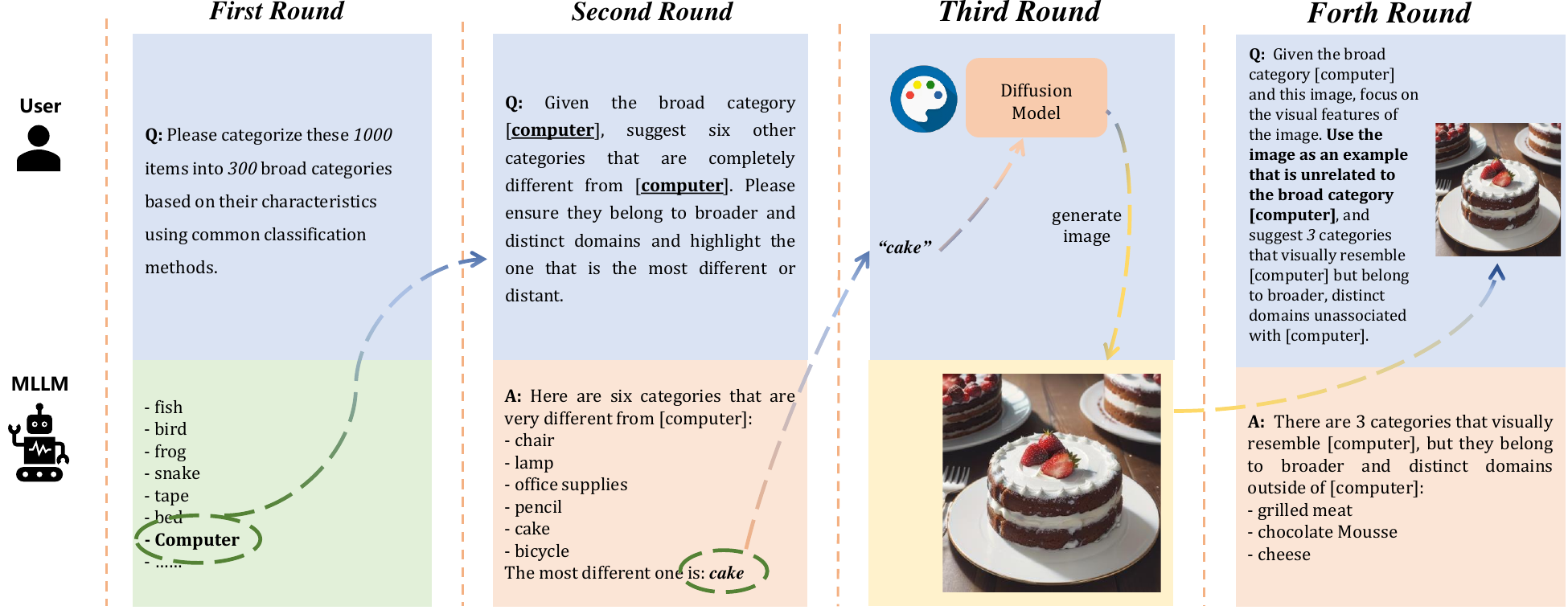}
\caption{Prompt design for multi-round conversation in the proposed MM-OOD framework under far out-of-distribution (OOD) detection.}
\label{fig:converation}
\end{figure*}
\noindent\textbf{MM-OOD Setups.}  
In our framework, we adopt CLIP~\cite{radford2021learning} as the core model, utilizing ViT-B/16 as the image encoder and a masked self-attention Transformer~\cite{vaswani2017attention} as the text encoder. For far OOD detection, we first leverage GPT-4 as the LLM to generate initial category labels. These labels serve as the foundation for generating OOD class labels. We then employ LLaVA (v1.5-7b) as the MLLM to propose candidate OOD labels. From this set of candidates, the OOD label most dissimilar to the primary category label is selected. To refine the label generation process, we introduce Stable Diffusion v1.5, which is used to generate an image. This generated image is then input back into LLaVA, guiding it to produce a final OOD label that is distinct from the primary category, ensuring that the generated outlier class is sufficiently differentiated.

\noindent\textbf{Baselines.}  
We compare our method against state-of-the-art zero-shot OOD detection approaches, all based on CLIP as the backbone. The experimental setup follows the configuration in EOE. Specifically, we evaluate our method alongside three key baselines in the zero-shot setting: MaxLogit~\cite{hendrycks2022scaling}, Energy~\cite{liu2020energy}, and MCM~\cite{ming2022delving}. These methods use the pre-trained CLIP model without fine-tuning, enabling a direct comparison with our approach.



\noindent\textbf{Evaluation Metrics.} To assess performance, we use two common evaluation metrics: (1) FPR95, which measures the false positive rate of OOD samples when the true positive rate for ID samples is fixed at 95\%. A lower FPR95 value indicates better detection capability; (2) AUROC, which calculates the area under the receiver operating characteristic curve, with a higher value reflecting better results. 
\begin{table*}[t]
\centering
\footnotesize 
\caption{Zero-shot far out-of-distribution (OOD) detection results on ImageNet-1K with varying numbers of primary categories $M$. Lower FPR95 and higher AUROC are better.}
\label{ood far L}
\setlength{\tabcolsep}{2.5pt} 
\begin{tabular}{lcccccccccccc}
\toprule
\multirow{2}{*}{\shortstack{\textbf{Primary Category} \\ \textbf{Number}}} &
\multirow{2}{*}{\textbf{Method}} &
\multicolumn{2}{c}{\textbf{iNaturalist}} &
\multicolumn{2}{c}{\textbf{SUN}} &
\multicolumn{2}{c}{\textbf{Places}} &
\multicolumn{2}{c}{\textbf{Texture}} &
\multicolumn{2}{c}{\textbf{Average}} \\
\cmidrule(lr){3-4} \cmidrule(lr){5-6} \cmidrule(lr){7-8} \cmidrule(lr){9-10} \cmidrule(lr){11-12}
& & \textbf{FPR95} $\downarrow$ & \textbf{AUROC} $\uparrow$ & \textbf{FPR95} $\downarrow$ & \textbf{AUROC} $\uparrow$ & \textbf{FPR95} $\downarrow$ & \textbf{AUROC} $\uparrow$ & \textbf{FPR95} $\downarrow$ & \textbf{AUROC} $\uparrow$ & \textbf{FPR95} $\downarrow$ & \textbf{AUROC} $\uparrow$ \\
\midrule
\multirow{2}{*}{$M=100$} 
& LLaVA-1.5 (EOE)   & 72.11 & 72.43 & 60.64 & 83.97 & 53.12 & 87.88 & 55.73 & 86.50 & 60.40 & 82.69 \\
& LLaVA-1.5 (Ours)  & 72.83 & 72.20 & 59.79 & 84.35 & 50.68 & 88.30 & 55.05 & 86.85 & 59.59 & 82.92 \\
\midrule
\multirow{2}{*}{$M=300$} 
& LLaVA-1.5 (EOE)   & 69.64 & 74.67 & 59.40 & 84.24 & 49.32 & 88.72 & 54.74 & 86.61 & 58.27 & 83.56 \\
& LLaVA-1.5 (Ours)  & 68.89 & 74.69 & 58.37 & 84.88 & 48.29 & 88.70 & 53.79 & 87.04 & 57.34 & 83.83 \\
\midrule
\multirow{2}{*}{$M=500$} 
& LLaVA-1.5 (EOE)   & 70.61 & 73.46 & 58.15 & 84.82 & 51.67 & 87.63 & 53.55 & 87.15 & 58.49 & 83.27 \\
& LLaVA-1.5 (Ours)  & 70.26 & 73.57 & 58.81 & 84.66 & 48.68 & 88.54 & 53.88 & 87.04 & 57.91 & 83.45 \\
\bottomrule
\end{tabular}
\end{table*}

\subsection{Main Results}
\label{exp_main}
\noindent\textbf{Near OOD Detection.}
Table~\ref{tab:near-ood} shows the performance of MM-OOD in near OOD detection on ImageNet-10 and ImageNet-20. For ImageNet-10, MM-OOD achieves an FPR95 of 3.84\%, improving by 3.17 percentage points compared to EOE, which has an FPR95 of 7.01\%, and by 9.97 percentage points compared to the Energy baseline, which has an FPR95 of 13.81\%. Similarly, for ImageNet-20, MM-OOD reduces FPR95 to 5.18\%, which is 1.19 percentage points lower than EOE’s 6.37\% and 2.81 percentage points lower than Energy’s 7.99\%. In summary, MM-OOD provides consistent improvements over both EOE and Energy across different settings in the near OOD detection task.


\noindent\textbf{Far OOD Detection.} 
Table~\ref{tab:far-ood} compares our method with state-of-the-art zero-shot OOD detection approaches across four ID datasets: CUB-200-2011, STANFORD-CARS, Food-101, and Oxford-IIIT Pet. The "Ground Truth” results, which assume access to the true OOD class labels, represent an ideal scenario but are not available in real-world applications, setting a theoretical upper bound. In average results, MM-OOD with $L=12 \times K$ outperforms all methods, achieving the best FPR95 of $4.33\%$ and AUROC of $99.56\%$, surpassing EOE, MaxLogit, and Energy. On individual datasets, MM-OOD also demonstrates strong performance, especially on CUB-200-2011, Food-101, and Oxford-IIIT Pet, where it achieves superior results compared to other methods. These results highlight the effectiveness ofMM-OOD in far OOD detection, showing consistent advantages in all evaluated scenarios.


\noindent\textbf{Scaling to ImageNet-1K dataset.} Table~\ref{ood far L} presents the performance of MM-OOD for far OOD detection on the ImageNet-1K dataset, comparing it to the baseline method EOE across different primary category numbers \( M \). This experiment is important for demonstrating that MM-OOD remains effective even when applied to large-scale, real-world datasets like ImageNet-1K. As shown in the table, MM-OOD achieves consistently better performance than EOE across all values of \( M \). Specifically, for \( M=100 \), MM-OOD improves FPR95 by 0.81\% and AUROC by 0.23\% compared to EOE. At \( M=300 \), MM-OOD shows improvements of 0.93\% in FPR95 and 0.27\% in AUROC. For \( M=500 \), MM-OOD further improves FPR95 by 0.58\% and AUROC by 0.18\%. Overall, MM-OOD shows a improvement, with an average increase of 0.77\% in FPR95 and 0.23\% in AUROC across all \( M \) values. These results suggest that MM-OOD is capable of handling large-scale far OOD detection tasks effectively, even as the number of primary categories increases.

\begin{table*}[t]
\centering
\caption{Zero-shot far out-of-distribution (OOD) detection results on Food-101 using LLaVA-1.5-7B and Qwen2-VL with varying numbers of primary categories $M$. The outlier label count is set to $12 \times K$, where $K$ denotes the number of in-distribution (ID) classes.}
\label{tab:model_variants}
\setlength{\tabcolsep}{2.5pt} 
\begin{tabular}{lcccccccccccc}
\toprule
\multirow{2}{*}{\shortstack{\textbf{Primary Category} \\ \textbf{Number}}} &
\multirow{2}{*}{\textbf{Method}} &
\multicolumn{2}{c}{\textbf{iNaturalist}} &
\multicolumn{2}{c}{\textbf{SUN}} &
\multicolumn{2}{c}{\textbf{Places}} &
\multicolumn{2}{c}{\textbf{Texture}} &
\multicolumn{2}{c}{\textbf{Average}} \\
\cmidrule(lr){3-4} \cmidrule(lr){5-6} \cmidrule(lr){7-8} \cmidrule(lr){9-10} \cmidrule(lr){11-12}
& & \textbf{FPR95} $\downarrow$ & \textbf{AUROC} $\uparrow$ & \textbf{FPR95} $\downarrow$ & \textbf{AUROC} $\uparrow$ & \textbf{FPR95} $\downarrow$ & \textbf{AUROC} $\uparrow$ & \textbf{FPR95} $\downarrow$ & \textbf{AUROC} $\uparrow$ & \textbf{FPR95} $\downarrow$ & \textbf{AUROC} $\uparrow$ \\
\midrule
\multirow{4}{*}{$M=10$}
& LLaVA-1.5 (EOE)    & 2.42 & 99.52 & 1.88 & 99.64 & 0.17 & 99.94 & 1.57 & 99.60 & 1.24 & 99.72 \\
& LLaVA-1.5 (Ours)   & 1.88 & 99.64 & 0.21 & 99.94 & 0.79 & 99.80 & 1.55 & 99.63 & 1.11 & 99.75 \\
& Qwen2-VL (EOE)     & 7.07 & 98.44 & 1.43 & 99.68 & 2.50 & 99.45 & 4.10 & 98.68 & 3.78 & 99.10 \\
& Qwen2-VL (Ours)    & 4.86 & 98.98 & 0.77 & 99.82 & 1.62 & 99.62 & 3.16 & 98.98 & 2.60 & 99.31 \\
\midrule
\multirow{4}{*}{$M=20$}
& LLaVA-1.5 (EOE)    & 4.62 & 99.00 & 0.66 & 99.82 & 1.33 & 99.67 & 2.27 & 99.50 & 2.22 & 99.50 \\
& LLaVA-1.5 (Ours)   & 2.00 & 99.60 & 0.28 & 99.92 & 0.86 & 99.79 & 1.34 & 99.69 & 1.12 & 99.75 \\
& Qwen2-VL (EOE)     & 7.50 & 98.32 & 1.74 & 99.62 & 2.97 & 99.35 & 4.65 & 98.56 & 4.22 & 98.96 \\
& Qwen2-VL (Ours)    & 6.57 & 98.51 & 1.26 & 99.71 & 2.30 & 99.47 & 4.45 & 98.53 & 3.65 & 99.10 \\
\bottomrule
\end{tabular}
\end{table*}

\begin{table*}[t]
\centering
\caption{Comparison results with random words in WordNet on far OOD detection task.}
\label{rebuttal:far-ood}
\setlength{\tabcolsep}{3.5pt}
\begin{tabular}{lcccccccccccc}
\toprule
\multirow{2}{*}{\textbf{ID Dataset}} &
\multirow{2}{*}{\textbf{Method}} &
\multicolumn{2}{c}{\textbf{iNaturalist}} &
\multicolumn{2}{c}{\textbf{SUN}} &
\multicolumn{2}{c}{\textbf{Places}} &
\multicolumn{2}{c}{\textbf{Texture}} &
\multicolumn{2}{c}{\textbf{Average}} \\
\cmidrule(lr){3-4} \cmidrule(lr){5-6} \cmidrule(lr){7-8} \cmidrule(lr){9-10} \cmidrule(lr){11-12}
& & \textbf{FPR95} $\downarrow$ & \textbf{AUROC} $\uparrow$ & \textbf{FPR95} $\downarrow$ & \textbf{AUROC} $\uparrow$ & \textbf{FPR95} $\downarrow$ & \textbf{AUROC} $\uparrow$ & \textbf{FPR95} $\downarrow$ & \textbf{AUROC} $\uparrow$ & \textbf{FPR95} $\downarrow$ & \textbf{AUROC} $\uparrow$ \\
\midrule
\multirow{2}{*}{CUB-200-2011}
& Ours + Random      & 16.72 & 94.37 & 0.04 & 99.99 & 0.42 & 99.96 & 0.02 & 99.96 & 4.30 & 98.57 \\
& Ours + LLaVA  &     
\cellcolor{gray!20}\textbf{15.23} & \cellcolor{gray!20}\textbf{94.86}  & \cellcolor{gray!20}\textbf{0.03} & \cellcolor{gray!20}\textbf{99.99} & \cellcolor{gray!20}\textbf{0.33}  & \cellcolor{gray!20}\textbf{99.98} & \cellcolor{gray!20}\textbf{0.02} & \cellcolor{gray!20}\textbf{99.99} & \cellcolor{gray!20}\textbf{3.90} & \cellcolor{gray!20}\textbf{98.69} \\ 
\bottomrule
\end{tabular}
\end{table*}

\begin{table}[t]
\centering
\footnotesize 
\setlength{\tabcolsep}{1pt} 
\caption{Zero-shot \textbf{near} OOD detection results. The \textbf{bold} indicates the best performance, and the gray color indicates the best performance on every dataset.}
\label{tab:near-ood}
\begin{tabular}{lcccccc}
\toprule
\multirow{2}{*}{\textbf{Method}} 
& \multicolumn{2}{c}{\textbf{ID}: ImageNet-10} 
& \multicolumn{2}{c}{\textbf{ID}: ImageNet-20} 
& \multicolumn{2}{c}{\textbf{Average}} \\
& \multicolumn{2}{c}{\textbf{OOD}: ImageNet-20} 
& \multicolumn{2}{c}{\textbf{OOD}: ImageNet-10} 
& \multicolumn{2}{c}{\phantom{AUROC$\uparrow$}} \\
\cmidrule(lr){2-3} \cmidrule(lr){4-5} \cmidrule(lr){6-7}
& \textbf{FPR95$\downarrow$} & \textbf{AUROC$\uparrow$} 
& \textbf{FPR95$\downarrow$} & \textbf{AUROC$\uparrow$} 
& \textbf{FPR95$\downarrow$} & \textbf{AUROC$\uparrow$} \\
\midrule
Energy    & 13.81 & 97.04 & 7.99 & 98.45 & 10.90 & 97.75 \\
MaxLogit  & 12.80 & 97.31 & 6.72 & 98.65 & 9.76  & 97.98 \\
MCM       & 5.72  & 98.61 & 9.05 & 98.39 & 7.39  & 98.50 \\
EOE       & 7.01  & 98.47 & 6.37 & 98.83 & 6.69  & 98.65 \\
\rowcolor{gray!20} \textbf{Ours}     & \textbf{3.84} & \textbf{98.97}  & \textbf{5.18} & \textbf{98.97} & \textbf{4.51} & \textbf{98.97} \\
Ground Truth & 0.20 & 99.80 & 0.20 & 99.93 & 0.20 & 99.87 \\
\bottomrule
\end{tabular}
\vspace{-0.2cm} 
\end{table}
\begin{table}[t]
\centering
\footnotesize
\setlength{\tabcolsep}{2pt} 
\caption{Performance and efficiency on different $L$.}
\label{tab11}
\begin{tabular}{lrrrrr}
\toprule
\multirow{2}{*}{Metrics} & \multicolumn{5}{c}{$L$} \\
        \cmidrule(lr){2-6}
         & 4$\times K$ & 6$\times K$ & 12$\times K$ & 40$\times K$ & 180$\times K$  \\
\midrule

EOE Time & 118m 36s & 120m 28s & 126m 14s & 189m 16s & 218m 34s \\
MM-OOD Time & 125m 8s & 124m 37s & 117m 28s & 184m 59s & 247m 55s \\
\midrule
EOE FPR95 & 2.14 & 2.13 & 1.07 & 0.80 & 1.25 \\
MM-OOD FPR95 & \cellcolor{gray!20}\textbf{1.63} & \cellcolor{gray!20}\textbf{1.39} & \cellcolor{gray!20}\textbf{0.45} & \cellcolor{gray!20}\textbf{0.75} & \cellcolor{gray!20}\textbf{0.67} \\
\midrule
EOE AUROC & 99.51 & 99.51 & 99.76 & 99.83 & 99.72 \\
MM-OOD AUROC & \cellcolor{gray!20}\textbf{99.64} & \cellcolor{gray!20}\textbf{99.69} & \cellcolor{gray!20}\textbf{99.90} & \cellcolor{gray!20}\textbf{99.84} & \cellcolor{gray!20}\textbf{99.86} \\
\bottomrule
\end{tabular}
\vspace{-0.2cm}
\end{table}

\noindent\textbf{Computational Efficiency.} Our method MM-OOD's computation time is similar to EOE. We conduct the experiments using the Food-101 dataset as the ID dataset as shown in Table ~\ref{tab11} on one A100 GPU. We conclude that there is little difference in time between EOE and MM-OOD. Additionally, as L increases, the performance of EOE improves, but MM-OOD shows even greater improvement. It shows that our method MM-OOD has scalability.



\subsection{Ablation Study}
We study the impact of outlier class number $L$, hyperparameter $\beta$, and different MLLMs (LLaVA, Qwen2-VL). Results confirm the robustness and effectiveness of our design.

\begin{figure}[t]
\centering
\includegraphics[width=0.7\linewidth]{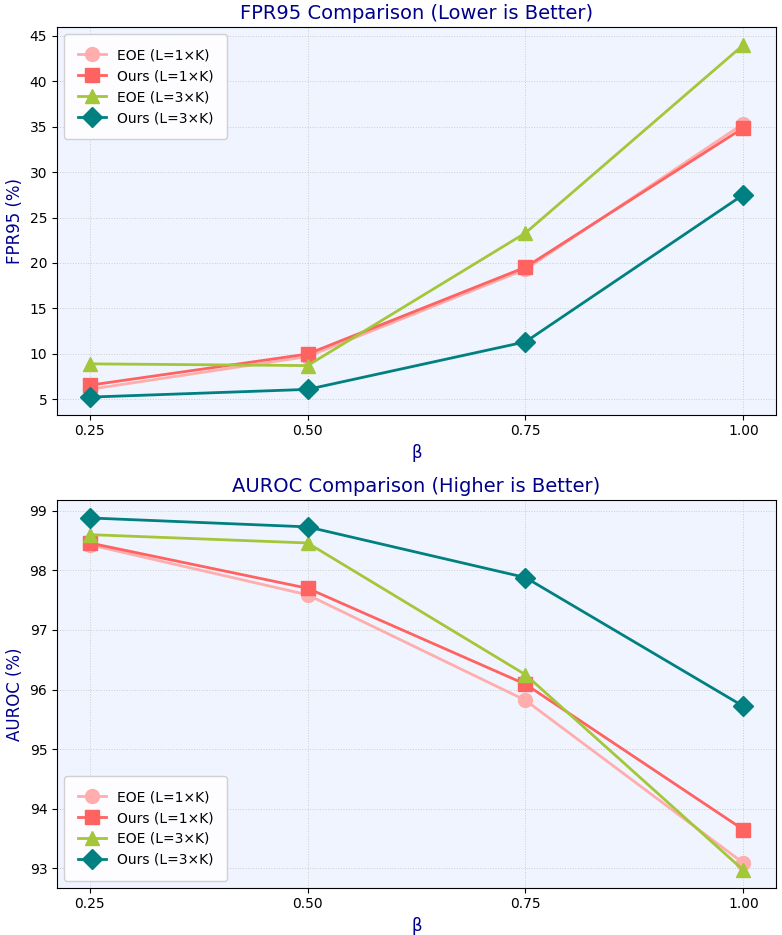}
\caption{Comparison of EOE and the proposed method under varying outlier class label counts $L = c \times K$ and threshold $\beta$ on the near out-of-distribution (OOD) detection task, where $K$ denotes the number of in-distribution (ID) classes.}
\label{fig:ablation}
\end{figure}

\noindent\textbf{Different MLLMs.} 
Table~\ref{Model Variants} presents the zero-shot far OOD detection performance of MM-OOD and EOE across two models: LLaVA-1.5 and Qwen2-VL. In all cases, MM-OOD achieves better results than EOE, demonstrating greater robustness in far OOD detection under various primary category settings. In addition, we also validate the results on InternVL2.5, which continue to demonstrate the superiority of EOE. For further details, please see Appendix.

\subsection{More Discussions}
\label{in_depth_discussion}
\noindent\textbf{The relationship between multi-round conversation and CoT.}
In our setup, $N$ is 1. In fact, it is only the parameter used for batch processing. Multi-round conversations refer to the sketch-generate-elaborate framework used for far-OOD detection tasks. In fact, sketch-generate-elaborate framework aligns with the Chain of Thought theory enhancing the model's instruction-following ability and accuracy through more detailed and iterative dialogues. Let's give an example to demonstrate the benefits mentioned above. When using the EOE method with the InternVL2.5 model, the output may be something like, "I can't understand the content of the image; the image you described includes...". In contrast, with our MM-OOD method, the OOD class result is accurately generated. 


\noindent\textbf{Unknown whether it is far OOD task or near OOD task at test time.} In some cases, it is not clear whether the OOD sample belongs to the far OOD task or near OOD task, making it impractical to select one of the branches of MM-OOD. To address this challenge, we propose proportionally mixing the OOD samples envisioned by both the near and far branches during the envisioning phase of the MLLM. A default mixing ratio of 0.5 can be used.

\noindent\textbf{Why not simply use random words on the far OOD task.} Based on the above, we know that MLLMs struggle to recommend far-OOD labels effectively. Therefore, one might question the feasibility of using random words instead. As shown in Table \ref{rebuttal:far-ood}, we designed the experiments and proved that MM-OOD can perform better than the random sampling from WordNet. We contend that this issue arises primarily because random sampling can indiscriminately select both near and far samples. Random samples tend to be dispersed and insufficiently concentrated, resulting in decreased performance and fluctuation in the far OOD task.






\section{Conclusion}
\label{conclusion}
We propose \textbf{MM-OOD}, a framework that leverages multimodal large language models (MLLMs) through multimodal reasoning and multi-round conversation to more efficiently obtain outlier samples, thereby improving out-of-distribution (OOD) detection performance. Specifically, by exploiting the inductive bias of MLLMs and synthesizing OOD images using generative models, our method effectively transforms near-OOD sampling into far-OOD sampling.


%

\bibliographystyle{IEEEtran}
\bibliography{ref}

@String{PAMI  = {IEEE Trans. Pattern Anal. Mach. Intell.}}

@String{CVPR  = {IEEE Conf. Comput. Vis. Pattern Recog.}}

@String{ICCV  = {Int. Conf. Comput. Vis.}}

@String{ECCV  = {Eur. Conf. Comput. Vis.}}

@String{NeurIPS = {Adv. Neural Inf. Process. Syst.}}

@String{ICLR  = {Int. Conf. Learn. Represent.}}

@String{IJCAI = {Int. Joint Conf. Artif. Intell.}}

@String{AAAI  = {AAAI Conf. Artif. Intell.}}

@inproceedings{chen2021atom,
  title     = {ATOM: Robustifying out-of-distribution detection using outlier mining},
  author    = {J. Chen and Y. Li and X. Wu and Y. Liang and S. Jha},
  booktitle = {Proc. Eur. Conf. Mach. Learn. Princ. Pract. Knowl. Discov. Databases (ECML PKDD)},
  year      = {2021}
}

@inproceedings{bai2024idlike,
  title     = {ID-like prompt learning for few-shot out-of-distribution detection},
  author    = {Y. Bai and Z. Han and C. Zhang and B. Cao and X. Jiang and Q. Hu},
  booktitle = CVPR,
  year      = {2024}
}

@inproceedings{cao2024envisioning,
  title     = {Envisioning outlier exposure by large language models for out-of-distribution detection},
  author    = {C. Cao and Z. Zhong and Z. Zhou and Y. Liu and T. Liu and B. Han},
  booktitle = {Int. Conf. Mach. Learn. (ICML)},
  year      = {2024}
}

@inproceedings{lin2021mood,
  title     = {MOOD: Multi-level out-of-distribution detection},
  author    = {Z. Lin and S. D. Roy and Y. Li},
  booktitle = CVPR,
  year      = {2021}
}

@article{wei2023chainofthoughtpromptingelicitsreasoning,
  title         = {Chain-of-thought prompting elicits reasoning in large language models},
  author        = {J. Wei and X. Wang and D. Schuurmans and M. Bosma and B. Ichter and F. Xia and E. Chi and Q. Le and D. Zhou},
  journal       = {arXiv preprint arXiv:2201.11903},
  year          = {2023},
 
}

@inproceedings{zhu2023unleashing,
  title     = {Unleashing mask: Explore the intrinsic out-of-distribution detection capability},
  author    = {J. Zhu and H. Li and J. Yao and T. Liu and J. Xu and B. Han},
  booktitle = {ICML},
  year      = {2023}
}

@inproceedings{wang2022watermarking,
  title     = {Watermarking for out-of-distribution detection},
  author    = {Q. Wang and F. Liu and Y. Zhang and J. Zhang and C. Gong and T. Liu and B. Han},
  booktitle = NeurIPS,
  year      = {2022}
}

@inproceedings{zhou2022rethinking,
  title     = {Rethinking reconstruction autoencoder-based out-of-distribution detection},
  author    = {Y. Zhou},
  booktitle = CVPR,
  year      = {2022}
}

@inproceedings{sun2022out,
  title     = {Out-of-distribution detection with deep nearest neighbors},
  author    = {Y. Sun and Y. Ming and X. Zhu and Y. Li},
  booktitle = {ICML},
  year      = {2022}
}

@inproceedings{sun2022dice,
  title     = {DICE: Leveraging sparsification for out-of-distribution detection},
  author    = {Y. Sun and Y. Li},
  booktitle = ECCV,
  year      = {2022}
}

@inproceedings{huang2021importance,
  title     = {On the importance of gradients for detecting distributional shifts in the wild},
  author    = {R. Huang and A. Geng and Y. Li},
  booktitle = NeurIPS,
  year      = {2021}
}

@inproceedings{liu2020energy,
  title     = {Energy-based out-of-distribution detection},
  author    = {W. Liu and X. Wang and J. Owens and Y. Li},
  booktitle = NeurIPS,
  year      = {2020}
}

@inproceedings{xiao2020likelihood,
  title     = {Likelihood regret: An out-of-distribution detection score for variational auto-encoder},
  author    = {Z. Xiao and Q. Yan and Y. Amit},
  booktitle = NeurIPS,
  year      = {2020}
}

@inproceedings{ren2019likelihood,
  title     = {Likelihood ratios for out-of-distribution detection},
  author    = {J. Ren and P. J. Liu and E. Fertig and J. Snoek and R. Poplin and M. Depristo and J. Dillon and B. Lakshminarayanan},
  booktitle = NeurIPS,
  year      = {2019}
}

@inproceedings{lee2018simple,
  title     = {A simple unified framework for detecting out-of-distribution samples and adversarial attacks},
  author    = {K. Lee and K. Lee and H. Lee and J. Shin},
  booktitle = NeurIPS,
  year      = {2018}
}

@inproceedings{hendrycks17baseline,
  title     = {A baseline for detecting misclassified and out-of-distribution examples in neural networks},
  author    = {D. Hendrycks and K. Gimpel},
  booktitle = ICLR,
  year      = {2017}
}

@inproceedings{zhou-etal-2021-contrastive,
  title     = {Contrastive out-of-distribution detection for pretrained transformers},
  author    = {W. Zhou and F. Liu and M. Chen},
  booktitle = {Proc. Conf. Empir. Methods Nat. Lang. Process. (EMNLP)},
  pages     = {1100--1111},
  year      = {2021},
  doi       = {10.18653/v1/2021.emnlp-main.84}
}

@article{lang2024a,
  title   = {A survey on out-of-distribution detection in NLP},
  author  = {H. Lang and Y. Zheng and Y. Li and J. Sun and F. Huang and Y. Li},
  journal = {Trans. Mach. Learn. Res.},
  year    = {2024},
  issn    = {2835-8856}
}

@article{OECC2021,
  title   = {Outlier exposure with confidence control for out-of-distribution detection},
  author  = {A.-A. Papadopoulos and M. R. Rajati and N. Shaikh and J. Wang},
  journal = {Neurocomputing},
  volume  = {441},
  pages   = {138--150},
  year    = {2021},
  doi     = {10.1016/j.neucom.2021.02.007}
}

@inproceedings{arora-etal-2021-types,
  title     = {Types of out-of-distribution texts and how to detect them},
  author    = {U. Arora and W. Huang and H. He},
  booktitle = {Proc. Conf. Empir. Methods Nat. Lang. Process. (EMNLP)},
  pages     = {10687--10701},
  year      = {2021},
  doi       = {10.18653/v1/2021.emnlp-main.835}
}

@inproceedings{likehood2021,
  title     = {Likelihood-free out-of-distribution detection with invertible generative models},
  author    = {A. Ahmadian and F. Lindsten},
  booktitle = {Proc. Int. Joint Conf. Artif. Intell. (IJCAI)},
  pages     = {2119--2125},
  year      = {2021},
  doi       = {10.24963/ijcai.2021/292}
}

@inproceedings{mcm2022,
  title     = {Delving into out-of-distribution detection with vision-language representations},
  author    = {Y. Ming and Z. Cai and J. Gu and Y. Sun and W. Li and Y. Li},
  booktitle = NeurIPS,
  year      = {2022}
}

@inproceedings{2023clipn,
  title     = {{CLIPN} for zero-shot OOD detection: Teaching {CLIP} to say no},
  author    = {H. Wang and Y. Li and H. Yao and X. Li},
  booktitle = ICCV,
  pages     = {1802--1812},
  year      = {2023}
}

@inproceedings{esmaeilpour2022zero,
  title     = {Zero-shot out-of-distribution detection based on the pre-trained model {CLIP}},
  author    = {S. Esmaeilpour and B. Liu and E. Robertson and L. Shu},
  booktitle = AAAI,
  year      = {2022}
}

@inproceedings{zhao2024lova3,
  title     = {{LOVA3}: Learning to visual question answering, asking and assessment},
  author    = {H. H. Zhao and P. Zhou and D. Gao and M. Z. Shou},
  booktitle = NeurIPS,
  year      = {2024}
}

@inproceedings{zhao2024genixer,
  title     = {{Genixer}: Empowering multimodal large language models as a powerful data generator},
  author    = {H. H. Zhao and P. Zhou and M. Z. Shou},
  booktitle = ECCV,
  year      = {2024}
}

@inproceedings{radford2021learning,
  title     = {Learning transferable visual models from natural language supervision},
  author    = {A. Radford and J. W. Kim and C. Hallacy and A. Ramesh and G. Goh and S. Agarwal and G. Sastry and A. Askell and P. Mishkin and J. Clark and G. Krueger and I. Sutskever},
  booktitle = {ICML},
  year      = {2021}
}

@inproceedings{li2022blip,
  title     = {{BLIP}: Bootstrapping language-image pre-training for unified vision-language understanding and generation},
  author    = {J. Li and D. Li and C. Xiong and S. Hoi},
  booktitle = {Proc. Int. Conf. Mach. Learn. (ICML)},
  pages     = {12888--12900},
  year      = {2022}
}

@inproceedings{alayrac2022flamingo,
  title     = {{Flamingo}: A visual language model for few-shot learning},
  author    = {J.-B. Alayrac and J. Donahue and P. Luc and A. Miech and I. Barr and Y. Hasson and K. Lenc and A. Mensch and K. Millican and M. Reynolds and R. Ring and E. Rutherford and S. Cabi and T. Han and Z. Gong and S. Samangooei and M. Monteiro and J. Menick and S. Borgeaud and A. Brock and A. Nematzadeh and S. Sharifzadeh and M. Binkowski and R. Barreira and O. Vinyals and A. Zisserman and K. Simonyan},
  booktitle = NeurIPS,
  year      = {2022}
}

@inproceedings{Mandal_2019_CVPR,
  title     = {Out-of-distribution detection for generalized zero-shot action recognition},
  author    = {D. Mandal and S. Narayan and S. K. Dwivedi and V. Gupta and S. Ahmed and F. S. Khan and L. Shao},
  booktitle = CVPR,
  year      = {2019}
}

@techreport{wah2011caltech,
  title       = {The Caltech-UCSD Birds-200-2011 dataset},
  author      = {C. Wah and S. Branson and P. Welinder and P. Perona and S. Belongie},
  institution = {California Institute of Technology},
  number      = {CNS-TR-2011-001},
  year        = {2011}
}

@inproceedings{krause20133d,
  title     = {3D object representations for fine-grained categorization},
  author    = {J. Krause and M. Stark and J. Deng and L. Fei-Fei},
  booktitle = {ICCV Workshops},
  year      = {2013}
}

@inproceedings{parkhi2012cats,
  title     = {Cats and dogs},
  author    = {O. M. Parkhi and A. Vedaldi and A. Zisserman and C. V. Jawahar},
  booktitle = CVPR,
  year      = {2012}
}

@inproceedings{deng2009imagenet,
  title     = {ImageNet: A large-scale hierarchical image database},
  author    = {J. Deng and W. Dong and R. Socher and L.-J. Li and K. Li and L. Fei-Fei},
  booktitle = CVPR,
  year      = {2009}
}

@inproceedings{van2018inaturalist,
  title     = {The iNaturalist species classification and detection dataset},
  author    = {G. Van Horn and O. Mac Aodha and Y. Song and Y. Cui and C. Sun and A. Shepard and H. Adam and P. Perona and S. Belongie},
  booktitle = CVPR,
  year      = {2018}
}

@inproceedings{xiao2010sun,
  title     = {SUN database: Large-scale scene recognition from abbey to zoo},
  author    = {J. Xiao and J. Hays and K. A. Ehinger and A. Oliva and A. Torralba},
  booktitle = CVPR,
  year      = {2010}
}

@article{zhou2017places,
  title   = {Places: A 10 million image database for scene recognition},
  author  = {B. Zhou and A. Lapedriza and A. Khosla and A. Oliva and A. Torralba},
  journal = PAMI,
  volume  = {40},
  number  = {6},
  pages   = {1452--1464},
  year    = {2017}
}

@inproceedings{cimpoi2014describing,
  title     = {Describing textures in the wild},
  author    = {M. Cimpoi and S. Maji and I. Kokkinos and S. Mohamed and A. Vedaldi},
  booktitle = CVPR,
  year      = {2014}
}

@inproceedings{huang2021mos,
  title     = {MOS: Towards scaling out-of-distribution detection for large semantic space},
  author    = {R. Huang and Y. Li},
  booktitle = CVPR,
  year      = {2021}
}

@inproceedings{ming2022delving,
  author    = {Y. Ming and Z. Cai and J. Gu and Y. Sun and W. Li and Y. Li},
  title     = {Delving into out-of-distribution detection with vision-language representations},
  booktitle = NeurIPS,
  year      = {2022}
}

@article{krizhevsky2009learning,
  title   = {Learning multiple layers of features from tiny images},
  author  = {A. Krizhevsky and G. Hinton},
  journal = {Master's thesis, Dept. Comput. Sci., Univ. Toronto},
  year    = {2009}
}

@inproceedings{vaswani2017attention,
  title     = {Attention is all you need},
  author    = {A. Vaswani and N. Shazeer and N. Parmar and J. Uszkoreit and L. Jones and A. N. Gomez and {\L}. Kaiser and I. Polosukhin},
  booktitle = NeurIPS,
  year      = {2017}
}

@article{yang2019deep,
  title={Deep learning for single image super-resolution: A brief review},
  author={Yang, Wenming and Zhang, Xuechen and Tian, Yapeng and Wang, Wei and Xue, Jing-Hao and Liao, Qingmin},
  journal={IEEE Transactions on Multimedia},
  volume={21},
  number={12},
  pages={3106--3121},
  year={2019},
  publisher={IEEE}
}

@inproceedings{hendrycks2022scaling,
  title     = {Scaling out-of-distribution detection for real-world settings},
  author    = {D. Hendrycks and S. Basart and M. Mazeika and A. Zou and J. Kwon and M. Mostajabi and J. Steinhardt and D. Song},
  booktitle = {ICML},
  year      = {2022}
}

@article{jin2022vwp,
  title={VWP: An efficient DRL-based autonomous driving model},
  author={Jin, Yan-Liang and Ji, Ze-Yu and Zeng, Dan and Zhang, Xiao-Ping},
  journal={IEEE Transactions on Multimedia},
  volume={26},
  pages={2096--2108},
  year={2022},
  publisher={IEEE}
}

@article{ma2023feature,
  title={Feature distribution representation learning based on knowledge transfer for long-tailed classification},
  author={Ma, Yanbiao and Jiao, Licheng and Liu, Fang and Yang, Shuyuan and Liu, Xu and Chen, Puhua},
  journal={IEEE Transactions on Multimedia},
  volume={26},
  pages={2772--2784},
  year={2023},
  publisher={IEEE}
}

@inproceedings{bossard14,
  author    = {L. Bossard and M. Guillaumin and L. Van Gool},
  title     = {Food-101 -- mining discriminative components with random forests},
  booktitle = {Proc. Eur. Conf. Comput. Vis. (ECCV)},
  pages     = {446--461},
  year      = {2014}
}

@inproceedings{sharifi2024,
  title     = {Gradient-regularized out-of-distribution detection},
  author    = {S. Sharifi and T. Entesari and B. Safaei and V. M. Patel and M. Fazlyab},
  booktitle = ECCV,
  year      = {2024}
}

@inproceedings{Tang_2024_CVPR,
  title     = {{CORES}: Convolutional response-based score for out-of-distribution detection},
  author    = {K. Tang and C. Hou and W. Peng and R. Chen and P. Zhu and W. Wang and Z. Tian},
  booktitle = CVPR,
  year      = {2024}
}

@article{min2017blind,
  title={Blind quality assessment based on pseudo-reference image},
  author={Min, Xiongkuo and Gu, Ke and Zhai, Guangtao and Liu, Jing and Yang, Xiaokang and Chen, Chang Wen},
  journal={IEEE Transactions on Multimedia},
  volume={20},
  number={8},
  pages={2049--2062},
  year={2017},
  publisher={IEEE}
}

@inproceedings{ho2024longtailed,
  title     = {Long-tailed anomaly detection with learnable class names},
  author    = {C.-H. Ho and K.-C. Peng and N. Vasconcelos},
  booktitle = CVPR,
  year      = {2024}
}

@article{zhu2023multi,
  title={Multi-modal structure-embedding graph transformer for visual commonsense reasoning},
  author={Zhu, Jian and Wang, Hanli and He, Bin},
  journal={IEEE Transactions on Multimedia},
  volume={26},
  pages={1295--1305},
  year={2023},
  publisher={IEEE}
}

@inproceedings{Yuan_2024Discrimin,
  title     = {Discriminability-driven channel selection for out-of-distribution detection},
  author    = {Y. Yuan and R. He and Y. Dong and Z. Han and Y. Yin},
  booktitle = CVPR,
  year      = {2024}
}

@misc{humblotrenaux2024noisy,
  title         = {A noisy elephant in the room: Is your out-of-distribution detector robust to label noise?},
  author        = {G. Humblot-Renaux and S. Escalera and T. B. Moeslund},
  booktitle     = CVPR,
  year          = {2024}
}

%

\section{Biography Section}
\appendices
\clearpage
\newpage
\clearpage

\section{More method details}
\label{sec:method details}

\noindent\textbf{Near OOD Detection prompt.} We use the following prompt to perform near Out-of-Distribution Detection.

'''

Q: Given the image category [husky dog] and this image, please suggest visually similar categories that are not directly related or belong to the same primary group as [husky dog]. Provide suggestions that share visual characteristics but are from broader and different domains than [husky dog].

A: There are 3 classes similar to [husky dog], and they are from broader and different domains than [husky dog]:

- gray wolf

- black stone

- red panda

Q: Given the image category [basketball], please suggest visually similar categories that are not directly related or belong to the same primary group as [basketball]. Provide suggestions that share visual characteristics but are from broader and different domains than [basketball].

A: There are 3 classes similar to [basketball], and they are from broader and different domains than [basketball]:

- balloons

- blowfish

- hat

Q: Given the image category [water jug], please suggest visually similar categories that are not directly related or belong to the same primary group as [water jug]. Provide suggestions that share visual characteristics but are from broader and different domains than [water jug].

A: There are 3 classes similar to [water jug], and they are from broader and different domains than [water jug]:

- trumpets

- helmets

- rucksacks

Q: Given the image category [{class\_info}] and this image, please suggest visually similar categories that are not directly related or belong to the same primary group as [{class\_info}]. Provide suggestions that share visual characteristics but are from broader and different domains than [{class\_info}].

A: There are {envision\_nums} classes similar to [{class\_info}], and they are from broader and different domains than [{class\_info}]:

'''

\section{Experimental Details}
\label{sec:experimental details}

\noindent\textbf{Near OOD Detection.} We use \texttt{ImageNet-10} and \texttt{ImageNet-20} as ID and OOD datasets in a complementary manner, where \texttt{ImageNet-10} serves as the ID dataset and \texttt{ImageNet-20} as the OOD dataset, and vice versa. Both datasets are subsets of \texttt{ImageNet-1K}. \texttt{ImageNet-10} consists of 10 classes designed to mimic the class distribution of \texttt{CIFAR-10}, which includes categories such as "dog", "cat", and "horse", among others. \texttt{ImageNet-20}, on the other hand, contains 20 classes that are semantically similar to those in \texttt{ImageNet-10}, expanding the variety of objects to include both animal and non-animal categories like "table" and "chair". These datasets were introduced by MCM to explore ID and OOD classification, where \texttt{ImageNet-10} is used for in-distribution tasks and \texttt{ImageNet-20} for out-of-distribution tasks, testing the model's ability to generalize across unseen classes.

\noindent\textbf{Far OOD Detection.} For ID datasets, we utilize several well-known benchmark datasets, including \texttt{Oxford-IIIT Pet}, \texttt{STANFORD-CARS}, \texttt{CUB-200-2011}, \texttt{Food-101}, and \texttt{ImageNet-1K}, each of which offers distinct characteristics and challenges for in-distribution classification. \texttt{Oxford-IIIT Pet} consists of 37 pet species, each with a relatively small number of images, focusing on fine-grained classification of animal breeds. \texttt{STANFORD-CARS} contains 196 car categories with diverse appearance variations, ideal for testing fine-grained recognition. \texttt{CUB-200-2011} is a bird species dataset with 200 categories, offering challenges due to high intra-class variance and inter-class similarity. \texttt{Food-101} includes 101 food categories, each containing 1,000 images, with the task of distinguishing between food items that are often visually similar. \texttt{ImageNet-1K}, a large-scale dataset with 1,000 object categories, is included to test the scalability of our method on more complex and diverse datasets. For OOD datasets, we employ large-scale collections that provide a wider range of unseen object types, including \texttt{Texture}, which contains 11,000 images from 11 texture categories, \texttt{Places}, which offers 400 scene categories, \texttt{iNaturalist}, a large-scale nature dataset with over 500,000 images from more than 8,000 species, and \texttt{SUN}, a scene recognition dataset with over 130 scene categories. Additionally, we use datasets curated by \texttt{MOS}, which provides a collection of out-of-distribution images for testing generalization across domains. Please refer to the Appendix for additional details on how each dataset is adapted for ID-OOD tasks.

\section{Algorithm flow}
The algorithmic process of the \textbf{\textit{sketch-generate-elaborate}} framework for far OOD detection tasks is detailed in Algorithm \ref{algo:far}. In our setup, $N$ is 1. In fact, this number is not very important, considering the limit on the number of tokens, it is just the parameter used for batch processing. 

\begin{algorithm}[tb]
    \caption{\textbf{\textit{sketch-generate-elaborate}} Algorithm in \textbf{MM-OOD} Framework for far OOD detection task}
    \label{algo:far}
    \renewcommand{\algorithmicrequire}{\textbf{Input:}}
    \renewcommand{\algorithmicensure}{\textbf{Output:}}
    \renewcommand{\algorithmicensure}{\textbf{Output:}}
	\begin{algorithmic}[1] 
            \REQUIRE ID class labels $\mathcal{Y}_{id}$, ID samples $\mathcal{X}_{id}$, MLLM, Generative model $G$, $N$ rounds;
		\ENSURE outlier class labels $\mathcal{Y}_{outlier}$
		\FOR {$r=1$ to $N$ rounds}
        \STATE {\textbf{\textit{sketch}} outlier class labels:\\
        $\mathcal{Y}_{sketch}$ = MLLM(prompt($\mathcal{Y}_{id}$))}
        \STATE {Select a representative label from $\mathcal{Y}_{id}$: \\
        $\mathcal{Y}_{representative}$ = MLLM(prompt($\mathcal{Y}_{sketch}$))}
        \STATE {\textbf{\textit{generate}} outlier sample:\\
        $\mathcal{X}_{ood}$ = $G$($\mathcal{Y}_{representative}$)}\\
        \STATE {\textbf{\textit{elaborate}} outlier class labels: \\
        $\mathcal{Y}_{outlier}$ = $\mathcal{Y}_{outlier}$ $\cup$ MLLM(prompt($\mathcal{Y}_{id} , \mathcal{X}_{ood}$))}
		\ENDFOR
        \STATE \textbf{return} $\mathcal{Y}_{outlier}$
	\end{algorithmic}
    \vspace{-0.1cm}
\end{algorithm}

\section{More experimental results}
\label{sec:experimental results}


The results in Table \ref{Model Variants} compare zero-shot out-of-distribution (OOD) detection performance for different models, specifically the \texttt{LLaVA-1.5} (7B) and \texttt{Qwen2-VL} models, using both the original \texttt{EOE} method and our proposed approach. The experiments were conducted on the Food101 dataset, with varying numbers of primary categories ($M$) from 10 to 40, and an Outlier Class Label Number set to $12 \times K$, where $K$ is the number of classes in the corresponding ID dataset. The performance metrics reported are FPR95 (False Positive Rate at 95\% True Positive Rate) and AUROC (Area Under the Receiver Operating Characteristic Curve). A lower FPR95 and a higher AUROC indicate better performance. For $M = 10$, our method (\texttt{LLaVA-1.5(Ours)}) achieves a notable improvement over the original \texttt{LLaVA-1.5(EOE)} with an FPR95 of 1.88 and AUROC of 99.64 on iNaturalist, compared to the EOE version which has an FPR95 of 2.42 and AUROC of 99.52. Similarly, \texttt{Qwen2-VL(Ours)} shows better performance with an FPR95 of 4.86 and AUROC of 98.98, surpassing the original \texttt{Qwen2-VL(EOE)} which has an FPR95 of 7.07 and AUROC of 98.44. For $M = 20$, \texttt{LLaVA-1.5(Ours)} again outperforms its EOE counterpart, achieving an FPR95 of 2.00 and AUROC of 99.60, compared to 4.62 and 99.00 for \texttt{LLaVA-1.5(EOE)}. \texttt{Qwen2-VL(Ours)} also shows improvements, with an FPR95 of 6.57 and AUROC of 98.51, outperforming \texttt{Qwen2-VL(EOE)} which has an FPR95 of 7.50 and AUROC of 98.32. For $M = 40$, \texttt{LLaVA-1.5(Ours)} achieves an FPR95 of 1.95 and AUROC of 99.62, improving upon the EOE version with FPR95 of 2.34 and AUROC of 99.52. Similarly, \texttt{Qwen2-VL(Ours)} delivers better results with an FPR95 of 4.50 and AUROC of 99.03, surpassing \texttt{Qwen2-VL(EOE)} which has an FPR95 of 5.14 and AUROC of 98.91. Finally, the \textbf{Average} row aggregates results across all datasets, showing that both \texttt{LLaVA-1.5(Ours)} and \texttt{Qwen2-VL(Ours)} outperform their respective EOE variants, with the average FPR95 and AUROC for \texttt{LLaVA-1.5(Ours)} being 1.94 and 99.62, respectively, and for \texttt{Qwen2-VL(Ours)} being 5.31 and 98.84. The addition of the Average row demonstrates the consistent improvement our method provides across different models and primary category configurations, further confirming the efficacy of our approach for OOD detection.

\begin{table*}[ht]
	\centering
  \caption{Zero-shot far OOD detection results with different model. Comparison of our method and the original EOE method on the LLaVA1.5-7b and Qwen2-VL models under different numbers of primary categories $M$ on the Food101 dataset, with an Outlier Class Label Number set to $12\times {K} $. $K$ is the number of classes in the corresponding ID dataset.}
    \resizebox{\linewidth}{!}{
	\begin{tabular}{lccccccccccccccc}
		\toprule
        &  &\multicolumn{8}{c}{\textbf{OOD Dataset}} & \multicolumn{2}{c}{\multirow{2}{*}{\textbf{Average}}}  \\
        \multirow{2}{*}{\shortstack{\textbf{Primary Category} \\ \textbf{Number}}} &\multirow{2}{*}{\textbf{Method}} & \multicolumn{2}{c}{iNaturalist} & \multicolumn{2}{c}{SUN} & \multicolumn{2}{c}{Places} & \multicolumn{2}{c}{Texture} & & &\\
          \cmidrule(r){3-4} \cmidrule(r){5-6} \cmidrule(r){7-8} \cmidrule(r){9-10} \cmidrule(r){11-12}  
		&  &\multicolumn{1}{c}{\textbf{FPR95}$\downarrow$} & \multicolumn{1}{c}{\textbf{AUROC}$\uparrow$} &
  \multicolumn{1}{c}{\textbf{FPR95}$\downarrow$} & 
  \multicolumn{1}{c}{\textbf{AUROC}$\uparrow$} & \multicolumn{1}{c}{\textbf{FPR95}$\downarrow$} &\multicolumn{1}{c}{\textbf{AUROC}$\uparrow$} & \multicolumn{1}{c}{\textbf{FPR95}$\downarrow$} & \multicolumn{1}{c}{\textbf{AUROC}$\uparrow$} & 
  \multicolumn{1}{c}{\textbf{FPR95}$\downarrow$} & \multicolumn{1}{c}{\textbf{AUROC}$\uparrow$}\\ 
		\cmidrule(r){1-1} \cmidrule(r){2-2}
  \cmidrule(r){3-4} \cmidrule(r){5-6} \cmidrule(r){7-8} \cmidrule(r){9-10} \cmidrule(r){11-12}  
		\multirow{4}{*}{\textbf{$M=10$}} 
  & LLaVA-1.5(EOE) & 2.42 & 99.52  & 1.88 & 99.64 & 0.17 & 99.94 & 1.57 & 99.6 & 1.24 & 99.72 \\
  & LLaVA-1.5(Ours) & 1.88 & 99.64  & 0.21 & 99.94 & 0.79 & 99.80 & 1.55 & 99.63 & 1.11 & 99.75 \\
  & Qwen2-VL(EOE) & 7.07 & 98.44  & 1.43 & 99.68 & 2.50  & 99.45 & 4.10 & 98.68 & 3.78 & 99.1 \\
 & Qwen2-VL(Ours) & 4.86 & 98.98  & 0.77 & 99.82 & 1.62  & 99.62 & 3.16 & 98.98 & 2.60 & 99.31 \\ 
        \midrule
        \multirow{4}{*}{\textbf{$M=20$}} 
  & LLaVA-1.5(EOE) & 4.62 & 99.00  & 0.66 & 99.82 & 1.33  & 99.67 & 2.27 & 99.50 & 2.22 & 99.50 \\
  & LLaVA-1.5(Ours) & 2.00 & 99.60  & 0.28 & 99.92 & 0.86  & 99.79 & 1.34 & 99.69 & 1.12 & 99.75 \\
  & Qwen2-VL(EOE) & 7.50 & 98.32  & 1.74 & 99.62 & 2.97  & 99.35 & 4.65 & 98.56 & 4.22 & 98.96 \\
 & Qwen2-VL(Ours) & 6.57 & 98.51  & 1.26 & 99.71 & 2.30  & 99.47 & 4.45 & 98.53 & 3.65 & 99.10 \\ 
        \midrule
        \multirow{4}{*}{\textbf{$M=40$}} 
  & LLaVA-1.5(EOE) & 2.34 & 99.52  & 0.21 & 99.94 & 0.81  & 99.80 & 1.53 & 99.63 & 1.22 & 99.72 \\
  & LLaVA-1.5(Ours) & 1.95 & 99.62  & 0.2 & 99.95 & 0.78  & 99.82 & 1.36 & 99.67 & 1.07 & 99.77 \\
  & Qwen2-VL(EOE) & 5.14 & 98.91  & 0.78 & 99.81 & 1.69  & 99.62 & 3.01 & 98.85 & 2.66 & 99.30 \\
 & Qwen2-VL(Ours) & 4.50 & 99.03  & 0.61 & 99.84 & 1.43  & 99.66 & 3.32 & 98.73 & 2.47 & 99.32 \\ 
        \midrule
        \multirow{4}{*}{\textbf{Average}} 
  & LLaVA-1.5(EOE) & 3.13 & 99.35  & 0.92 & 99.80 & 0.77  & 99.80 & 1.79 & 99.58 & 	1.65 & 99.63 \\
  & \cellcolor{gray!20}\textbf{LLaVA-1.5(Ours)} & \cellcolor{gray!20}\textbf{1.94} & \cellcolor{gray!20}\textbf{99.62} & \cellcolor{gray!20}\textbf{0.23} & \cellcolor{gray!20}\textbf{99.94} & \cellcolor{gray!20}\textbf{0.81} & \cellcolor{gray!20}\textbf{99.80} & \cellcolor{gray!20}\textbf{1.42} & \cellcolor{gray!20}\textbf{99.66} & \cellcolor{gray!20}\textbf{1.10} & \cellcolor{gray!20}\textbf{99.76} \\
  & Qwen2-VL(EOE) & 6.57 & 98.56  & 1.32 & 99.70 & 2.39  & 99.47 & 3.92 & 98.70 & 3.55 & 99.11 \\
  & \cellcolor{gray!20}\textbf{Qwen2-VL(Ours)} & \cellcolor{gray!20}\textbf{5.31} & \cellcolor{gray!20}\textbf{98.84} & \cellcolor{gray!20}\textbf{0.88} & \cellcolor{gray!20}\textbf{99.79} & \cellcolor{gray!20}\textbf{1.78} & \cellcolor{gray!20}\textbf{99.58} & \cellcolor{gray!20}\textbf{3.64} & \cellcolor{gray!20}\textbf{98.75} & \cellcolor{gray!20}\textbf{2.90} & \cellcolor{gray!20}\textbf{99.24} \\
		\bottomrule
	\end{tabular}}
   
     \label{Model Variants}
\end{table*}

The results presented in Table \ref{ood far L} showcase zero-shot out-of-distribution (OOD) detection performance for the \texttt{LLaVA-1.5} model, comparing the original \texttt{EOE} method with our proposed approach across different values of the primary category number ($M$) on the ImageNet-1K dataset. The performance is evaluated using two key metrics: FPR95 (False Positive Rate at 95\% True Positive Rate) and AUROC (Area Under the Receiver Operating Characteristic Curve), with lower FPR95 and higher AUROC indicating better performance. For each value of $M$, results are reported across four OOD datasets: iNaturalist, SUN, Places, and Texture. In terms of FPR95, our method (denoted as \texttt{LLaVA-1.5(Ours)}) shows marginal improvement or maintains competitive performance compared to the \texttt{EOE} method. For $M = 100$, the FPR95 of \texttt{LLaVA-1.5(Ours)} is slightly higher (72.83) than \texttt{LLaVA-1.5(EOE)} (72.11), but the AUROC is slightly lower for \texttt{LLaVA-1.5(Ours)} (72.20) compared to \texttt{LLaVA-1.5(EOE)} (72.43). For $M = 300$, \texttt{LLaVA-1.5(Ours)} achieves similar results with an FPR95 of 68.89 and AUROC of 74.69, compared to \texttt{LLaVA-1.5(EOE)} with FPR95 of 69.64 and AUROC of 74.67. For $M = 500$, the difference is even smaller, with \texttt{LLaVA-1.5(Ours)} achieving an FPR95 of 70.26 and AUROC of 73.57, compared to \texttt{LLaVA-1.5(EOE)}'s FPR95 of 70.61 and AUROC of 73.46. The average row shows that the performance of both methods is close across all values of $M$, with the average FPR95 for \texttt{LLaVA-1.5(Ours)} being 70.66 and AUROC of 73.49, while \texttt{LLaVA-1.5(EOE)} has an average FPR95 of 70.79 and AUROC of 73.52. Although the performance differences are small, \texttt{LLaVA-1.5(Ours)} demonstrates a slightly better overall performance, especially in terms of AUROC. The addition of the "Average" row emphasizes that, although there are minor fluctuations, our method maintains comparable or better performance across different numbers of primary categories and OOD datasets, confirming the robustness of our approach for zero-shot OOD detection tasks.

\begin{table*}[ht]
    \centering
    \caption{Zero-shot far OOD detection results with different Primary Category Numbers $M$ on ImageNet-1K.}
    \resizebox{\linewidth}{!}{
    \begin{tabular}{lccccccccccccccc}
	\toprule
        &  &\multicolumn{8}{c}{\textbf{OOD Dataset}} & \multicolumn{2}{c}{\multirow{2}{*}{\textbf{Average}}}  \\
        \multirow{2}{*}{\shortstack{\textbf{Primary Category} \\ \textbf{Number}}} &\multirow{2}{*}{\textbf{Method}} & \multicolumn{2}{c}{iNaturalist} & \multicolumn{2}{c}{SUN} & \multicolumn{2}{c}{Places} & \multicolumn{2}{c}{Texture} & & &\\
          \cmidrule(r){3-4} \cmidrule(r){5-6} \cmidrule(r){7-8} \cmidrule(r){9-10} \cmidrule(r){11-12}  
		&  &\multicolumn{1}{c}{\textbf{FPR95}$\downarrow$} & \multicolumn{1}{c}{\textbf{AUROC}$\uparrow$} &
  \multicolumn{1}{c}{\textbf{FPR95}$\downarrow$} & 
  \multicolumn{1}{c}{\textbf{AUROC}$\uparrow$} & \multicolumn{1}{c}{\textbf{FPR95}$\downarrow$} &\multicolumn{1}{c}{\textbf{AUROC}$\uparrow$} & \multicolumn{1}{c}{\textbf{FPR95}$\downarrow$} & \multicolumn{1}{c}{\textbf{AUROC}$\uparrow$} & 
  \multicolumn{1}{c}{\textbf{FPR95}$\downarrow$} & \multicolumn{1}{c}{\textbf{AUROC}$\uparrow$}\\ 
		\cmidrule(r){1-1} \cmidrule(r){2-2}
  \cmidrule(r){3-4} \cmidrule(r){5-6} \cmidrule(r){7-8} \cmidrule(r){9-10} \cmidrule(r){11-12}  
		\multirow{2}{*}{\textbf{$M=100$}} 
  & LLaVA-1.5(EOE) & 72.11 & 72.43  & 60.64 & 83.97 & 53.12 & 87.88 & 55.73 & 86.50 & 60.40 & 82.69 \\
  & LLaVA-1.5(Ours) & 72.83 & 72.20  & 59.79 & 84.35 & 50.68 & 88.30 & 55.05 & 86.85 & 59.59 & 82.92 \\
        \midrule
        \multirow{2}{*}{\textbf{$M=300$}} 
  & LLaVA-1.5(EOE) & 69.64 & 74.67  & 59.4 & 84.24 &  49.32 & 88.72 & 54.74 & 86.61 & 58.27 & 83.56 \\
  & LLaVA-1.5(Ours) & 68.89 & 74.69 & 58.37 & 84.88 & 48.29 & 88.7 & 53.79 & 87.04 & 57.34 & 83.83 \\
        \midrule
        \multirow{2}{*}{\textbf{$M=500$}} 
  & LLaVA-1.5(EOE) & 70.61 & 73.46 & 58.15 & 84.82 & 51.67 & 87.63 & 53.55 & 87.15 & 58.49 & 83.27 \\
  & LLaVA-1.5(Ours) & 70.26 & 73.57 & 58.81 & 84.66 &  48.68 & 88.54 & 53.88 & 87.04 & 57.91 & 83.45 \\
        \midrule
        \multirow{2}{*}{\textbf{Average}} 
  & LLaVA-1.5(EOE) & 70.79 & 73.52 & 59.40 & 84.34 & 51.37 & 88.08 & 54.67 & 86.75 & 59.06 & 83.17 \\
  & \cellcolor{gray!20}\textbf{LLaVA-1.5(Ours)} & \cellcolor{gray!20}70.66 & \cellcolor{gray!20}73.49 & \cellcolor{gray!20}58.99 & \cellcolor{gray!20}84.63 & \cellcolor{gray!20}49.22 & \cellcolor{gray!20}88.51 & \cellcolor{gray!20}54.24 & \cellcolor{gray!20}86.98 & \cellcolor{gray!20}\textbf{58.28} & \cellcolor{gray!20}\textbf{83.40} \\
        \bottomrule
    \end{tabular}}
    
     \label{ood far L}
\end{table*}

The results in Table \ref{ood far dataset & N} summarize the zero-shot far OOD detection performance across different ID datasets, varying broad category numbers ($N$), and OOD datasets. The performance is measured using FPR95 (False Positive Rate at 95\% True Positive Rate) and AUROC (Area Under the Receiver Operating Characteristic Curve), with lower FPR95 and higher AUROC values indicating better performance. The analysis highlights the comparison between the \texttt{EOE} method and our proposed approach (\texttt{Ours}). For the \textbf{CUB-200-2011} dataset, as $N$ increases from 10 to 40, our method shows consistent improvements over \texttt{EOE}, especially in AUROC, achieving a maximum of 99.97 at $N=40$ with FPR95 of 0.08 for the Texture OOD dataset. Similarly, for \textbf{STANFORD-CARS}, both methods perform nearly identically across all $N$ values, consistently delivering high AUROC scores close to 100. However, our method maintains a slight edge with marginally lower FPR95 for some datasets. On the \textbf{Food-101} dataset, our approach outperforms \texttt{EOE} at all $N$ values, particularly at $N=40$, where it achieves an average AUROC of 99.69 compared to 99.66 for \texttt{EOE}. For the \textbf{Oxford-IIIT Pet} dataset, \texttt{Ours} demonstrates significant improvement, particularly at $N=10$ and $N=20$, where it achieves AUROC values of 99.89 and 99.92, respectively, compared to 99.72 and 99.84 for \texttt{EOE}, alongside lower FPR95. The \textbf{Average} row consolidates these findings, revealing that across all datasets and $N$ values, our method consistently achieves better average performance, with an AUROC of 99.46 at $N=10$, 99.48 at $N=20$, and 99.73 at $N=40$, compared to 99.37, 99.42, and 99.41 for \texttt{EOE}. The addition of the "Average" row highlights the robustness and generalizability of our method across diverse datasets and configurations, demonstrating its superior performance in both fine-grained and large-scale OOD detection tasks.

\begin{table*}[ht]
	\centering
 \caption{Zero-shot far OOD detection results with different broad category number $N$.}
    \resizebox{\linewidth}{!}{
	\begin{tabular}{lccccccccccccccccc}
		\toprule
        &  &  & \multicolumn{8}{c}{\textbf{OOD Dataset}} & \multicolumn{2}{c}{\multirow{2}{*}{\textbf{Average}}}  \\
        \multirow{2}{*}{\textbf{ID Datasets}} 
        & \multirow{2}{*}{\shortstack{\textbf{Broad Category} \\ \textbf{Number}}}
        & \multirow{2}{*}{\textbf{Method}} 
        & \multicolumn{2}{c}{iNaturalist} 
        & \multicolumn{2}{c}{SUN} 
        & \multicolumn{2}{c}{Places} 
        & \multicolumn{2}{c}{Texture} 
        & & &\\
          \cmidrule(r){4-5} \cmidrule(r){6-7} \cmidrule(r){8-9} \cmidrule(r){10-11} \cmidrule(r){12-13}  
		&  & & \multicolumn{1}{c}{\textbf{FPR95}$\downarrow$} 
        & \multicolumn{1}{c}{\textbf{AUROC}$\uparrow$} 
        & \multicolumn{1}{c}{\textbf{FPR95}$\downarrow$} 
        & \multicolumn{1}{c}{\textbf{AUROC}$\uparrow$} 
        & \multicolumn{1}{c}{\textbf{FPR95}$\downarrow$} 
        & \multicolumn{1}{c}{\textbf{AUROC}$\uparrow$} 
        & \multicolumn{1}{c}{\textbf{FPR95}$\downarrow$} 
        & \multicolumn{1}{c}{\textbf{AUROC}$\uparrow$} 
        & \multicolumn{1}{c}{\textbf{FPR95}$\downarrow$} 
        & \multicolumn{1}{c}{\textbf{AUROC}$\uparrow$}\\ 
		\cmidrule(r){1-1} \cmidrule(r){2-2} \cmidrule(r){3-3} 
        \cmidrule(r){4-5} \cmidrule(r){6-7} \cmidrule(r){8-9} \cmidrule(r){10-11} \cmidrule(r){12-13}  
		\multirow{6}{*}{\textbf{CUB-200-2011}} 
    & N=10 & EOE & 17.62 & 94.21 & 1.10 & 99.71 & 2.14 & 99.53 & 0.62 & 99.85 & 5.37 & 98.32 \\
  & N=10 & Ours & 17.16 & 94.4 & 1.12 & 99.72 & 1.91 & 99.58 & 0.87 & 99.81 & 5.27 & 98.38 \\
  & N=20 & EOE & 17.01 & 94.39  & 1.19 & 99.71 & 2.01  & 99.56 & 0.70 & 99.85 & 5.23 & 98.38 \\
  & N=20 & Ours & 17.14 & 94.35  & 0.94 & 99.76 & 1.77 & 99.63 & 0.43 & 99.90 & 5.07 & 98.41 \\
  & N=40 & EOE & 15.88 & 94.74  & 0.40 & 99.88 & 0.73 & 99.83 & 0.17 & 99.95 & 4.29 & 98.60 \\
  & N=40 & Ours & 15.77 & 98.44  & 0.31 & 99.90 & 0.67 & 99.84 & 0.08 & 99.97 & 4.21 & 98.61 \\
        \midrule
        \multirow{6}{*}{\textbf{STANFORD-CARS}} 
    & N=10 & EOE & 0.02 & 99.98 & 0.08 & 99.98 & 0.37  & 99.90 & 0.00 & 100.00 & 0.12 & 99.97 \\
  & N=10 & Ours & 0.01 & 99.98  & 0.10 & 99.97 & 0.38 & 99.90 & 0.00 & 100.00 & 0.12 & 99.96 \\
  & N=20 & EOE & 0.00 & 99.99  & 0.09 & 99.98 & 0.36 & 99.90 & 0.00 & 100.00 & 0.11 & 99.97 \\
    & N=20 & Ours & 0.00 & 99.99  & 0.08 & 99.98 & 0.36 & 99.90 & 0.00 & 100.00 & 0.11 & 99.97 \\
  & N=40 & EOE & 0.00 & 100.00  & 0.10 & 99.98 & 0.35 & 99.91 & 0.00 & 100.00 & 0.11 & 99.97 \\
  & N=40 & Ours & 0.00 & 100.00  & 0.09 & 99.98 & 0.35 & 99.91 & 0.00 & 100.00 & 0.11 & 99.97 \\
        \midrule
        \multirow{6}{*}{\textbf{Food-101}} 
    & N=10 & EOE & 5.36 & 98.81 & 0.73 & 99.80 & 1.44 & 99.64 & 2.00 & 99.50 & 2.38 & 99.44 \\
  & N=10 & Ours & 3.56 & 99.30 & 0.54 & 99.87 & 1.25 & 99.71 & 1.90 & 99.53 & 1.81 & 99.60 \\
  & N=20 & EOE & 5.12 & 98.88  & 0.52 & 99.85 & 1.22  & 99.69 & 2.09 & 99.50 & 2.24 & 99.48 \\
  & N=20 & Ours & 3.59 & 99.26  & 0.46 & 99.88 & 0.94 & 99.77 & 1.84 & 99.57 & 1.71 & 99.62 \\
  & N=40 & EOE & 3.12 & 99.36  & 0.23 & 99.93 & 0.88 & 99.78 & 1.94 & 99.56 & 2.87 & 99.66 \\
  & N=40 & Ours & 2.85 & 99.43  & 0.27 & 99.92 & 0.89 & 99.78 & 1.55 & 99.64 & 1.39 & 99.69 \\
        \midrule
        \multirow{6}{*}{\textbf{Oxford-IIIT Pet}} 
    & N=10 & EOE & 0.55 & 99.79 & 1.15 & 99.72 & 1.85  & 99.52 & 0.52 & 99.84 & 1.02 & 99.72 \\
  & N=10 & Ours & 0.07 & 99.95  & 0.43 & 99.89 & 0.84 & 99.78 & 0.14 & 99.93 & 0.37 & 99.89 \\
  & N=20 & EOE & 0.21 & 99.89  & 0.6 & 99.86 & 1.13 & 99.71 & 0.29 & 99.89 & 0.56 & 99.84 \\
  & N=20 & Ours & 0.04 & 99.98  & 0.28 & 99.93 & 0.67 & 99.83 & 0.14 & 99.96 & 0.28 & 99.92 \\
  & N=40 & EOE & - & - & - & - & - & - & - & - & - & - \\
  & N=40 & Ours & - & - & - & - & -  & - & - & - & - & - \\
        \midrule
        \multirow{6}{*}{\textbf{Average}}
    & N=10 & EOE & 5.89 & 98.20 & 0.77 & 99.81 & 1.45 & 99.65 & 0.79 & 99.80 & 2.23 & 99.37 \\
    & N=10 & Ours & 5.20 & 98.41 & 0.55 & 99.86 & 1.10 & 99.74 & 0.73 & 99.82 & 1.89 & 99.46\\
    & N=20 & EOE & 5.59 & 98.29 & 0.60 & 99.85 & 1.18 & 99.72 & 0.77 & 99.81 & 2.04 & 99.42\\
    & N=20 & Ours & 5.19 & 98.4 & 0.44 & 99.89 & 0.94 & 99.78 & 0.60 & 99.86 & 1.79 & 99.48\\
    & N=40 & EOE & 6.33 & 98.03 & 0.24 & 99.93 & 0.65 & 99.84 & 0.70 & 99.84  & 1.98 & 99.41\\
    & N=40 & Ours & 6.21 & 99.29 & 0.22 & 99.93 & 0.64 & 99.84 & 0.54 & 99.87  & 1.90 & 99.73\\
		\bottomrule
	\end{tabular}}
    
     \label{ood far dataset & N}
\end{table*}

Table~\ref{result1} presents zero-shot far OOD detection results using Food101 as the ID dataset, with evaluations conducted under varying broad category numbers ($N$) and weighting parameters ($\beta$). The comparison is between two methods: \texttt{EOE} and \texttt{MM-OOD}, across four OOD datasets (iNaturalist, SUN, Places, and Texture), with the Outlier Class Label Number fixed at $L = 6 \times K$. The performance metrics are FPR95 (False Positive Rate at 95\% True Positive Rate) and AUROC (Area Under the Receiver Operating Characteristic Curve), where lower FPR95 and higher AUROC values indicate superior performance. For $N = 10$, \texttt{MM-OOD} consistently outperforms \texttt{EOE} across all $\beta$ values, particularly at $\beta = 0.25$, where it achieves an average FPR95 of 1.81 and AUROC of 99.60, compared to \texttt{EOE}'s 2.38 and 99.44, respectively. As $N$ increases to 20, \texttt{MM-OOD} maintains its advantage, with a notable improvement at $\beta = 0.25$, where it achieves an average FPR95 of 1.29 and AUROC of 99.70, demonstrating better generalization to OOD samples. For $N = 40$, \texttt{MM-OOD} again shows robust performance across all $\beta$ values, achieving an AUROC of 99.69 at $\beta = 0.25$, compared to \texttt{EOE}'s 99.66, while maintaining a slightly lower average FPR95 of 1.39. As $\beta$ increases to 0.75, both methods experience performance degradation, but \texttt{MM-OOD} consistently performs better than \texttt{EOE}, particularly at higher $N$ values. The Average row highlights the overall superiority of \texttt{MM-OOD}.

\noindent\textbf{Primary Category Number ($M$) in Far OOD Detection.} Table \ref{ood far dataset & N} presents the zero-shot far OOD detection performance of our method compared to EOE across different values of primary category number \( M \) (10 and 20). At \( M=10 \), our method achieves an average FPR95 of 1.89\% and AUROC of 99.46\%, outperforming EOE’s 2.23\% and 99.37\%, respectively. When \( M=20 \), the performance difference remains evident, with our method achieving an average FPR95 of 1.79\% and AUROC of 99.48\%, compared to EOE’s 2.04\% and 99.42\%. The improvement is particularly notable in challenging cases, such as the CUB-200-2011 dataset, where our method reduces FPR95 from 5.37\% (EOE) to 5.07\% and increases AUROC from 98.32\% to 98.41\%. Overall, our approach outperforms EOE across various datasets and primary category configurations.

\begin{table*}[ht]
	\centering
 \caption{Zero-shot far OOD detection results with different primary category number $M$. The \textbf{bold} indicates the best performance, and the gray color indicates the best performance on every dataset.}
    \resizebox{\linewidth}{!}{
	\begin{tabular}{lccccccccccccccccc}
		\toprule
        &  &  & \multicolumn{8}{c}{\textbf{OOD Dataset}} & \multicolumn{2}{c}{\multirow{2}{*}{\textbf{Average}}}  \\
        \multirow{2}{*}{\textbf{ID Datasets}} 
        & \multirow{2}{*}{\shortstack{\textbf{Primary Category} \\ \textbf{Number}}}
        & \multirow{2}{*}{\textbf{Method}} 
        & \multicolumn{2}{c}{iNaturalist} 
        & \multicolumn{2}{c}{SUN} 
        & \multicolumn{2}{c}{Places} 
        & \multicolumn{2}{c}{Texture} 
        & & &\\
          \cmidrule(r){4-5} \cmidrule(r){6-7} \cmidrule(r){8-9} \cmidrule(r){10-11} \cmidrule(r){12-13}  
		&  & & \multicolumn{1}{c}{\textbf{FPR95}$\downarrow$} 
        & \multicolumn{1}{c}{\textbf{AUROC}$\uparrow$} 
        & \multicolumn{1}{c}{\textbf{FPR95}$\downarrow$} 
        & \multicolumn{1}{c}{\textbf{AUROC}$\uparrow$} 
        & \multicolumn{1}{c}{\textbf{FPR95}$\downarrow$} 
        & \multicolumn{1}{c}{\textbf{AUROC}$\uparrow$} 
        & \multicolumn{1}{c}{\textbf{FPR95}$\downarrow$} 
        & \multicolumn{1}{c}{\textbf{AUROC}$\uparrow$} 
        & \multicolumn{1}{c}{\textbf{FPR95}$\downarrow$} 
        & \multicolumn{1}{c}{\textbf{AUROC}$\uparrow$}\\ 
		\cmidrule(r){1-1} \cmidrule(r){2-2} \cmidrule(r){3-3} 
        \cmidrule(r){4-5} \cmidrule(r){6-7} \cmidrule(r){8-9} \cmidrule(r){10-11} \cmidrule(r){12-13}  
		\multirow{4}{*}{\textbf{CUB-200-2011}} 
    & $M=10$ & EOE & 17.62 & 94.21 & 1.10 & 99.71 & 2.14 & 99.53 & 0.62 & 99.85 & 5.37 & 98.32 \\
  & $M=10$ & Ours & 17.16 & 94.4 & 1.12 & 99.72 & 1.91 & 99.58 & 0.87 & 99.81 & 5.27 & 98.38 \\
  & $M=20$ & EOE & 17.01 & 94.39  & 1.19 & 99.71 & 2.01  & 99.56 & 0.70 & 99.85 & 5.23 & 98.38 \\
  & $M=20$ & Ours & 17.14 & 94.35  & 0.94 & 99.76 & 1.77 & 99.63 & 0.43 & 99.90 & 5.07 & 98.41 \\
        \midrule
        \multirow{4}{*}{\textbf{STANFORD-CARS}} 
    & $M=10$ & EOE & 0.02 & 99.98 & 0.08 & 99.98 & 0.37  & 99.90 & 0.00 & 100.00 & 0.12 & 99.97 \\
  & $M=10$ & Ours & 0.01 & 99.98  & 0.10 & 99.97 & 0.38 & 99.90 & 0.00 & 100.00 & 0.12 & 99.96 \\
  & $M=20$ & EOE & 0.00 & 99.99  & 0.09 & 99.98 & 0.36 & 99.90 & 0.00 & 100.00 & 0.11 & 99.97 \\
    & $M=20$ & Ours & 0.00 & 99.99  & 0.08 & 99.98 & 0.36 & 99.90 & 0.00 & 100.00 & 0.11 & 99.97 \\
        \midrule
        \multirow{4}{*}{\textbf{Food-101}} 
    & $M=10$ & EOE & 5.36 & 98.81 & 0.73 & 99.80 & 1.44 & 99.64 & 2.00 & 99.50 & 2.38 & 99.44 \\
  & $M=10$ & Ours & 3.56 & 99.30 & 0.54 & 99.87 & 1.25 & 99.71 & 1.90 & 99.53 & 1.81 & 99.60 \\
  & $M=20$ & EOE & 5.12 & 98.88  & 0.52 & 99.85 & 1.22  & 99.69 & 2.09 & 99.50 & 2.24 & 99.48 \\
  & $M=20$ & Ours & 3.59 & 99.26  & 0.46 & 99.88 & 0.94 & 99.77 & 1.84 & 99.57 & 1.71 & 99.62 \\
        \midrule
        \multirow{4}{*}{\textbf{Oxford-IIIT Pet}} 
    & $M=10$ & EOE & 0.55 & 99.79 & 1.15 & 99.72 & 1.85  & 99.52 & 0.52 & 99.84 & 1.02 & 99.72 \\
  & $M=10$ & Ours & 0.07 & 99.95  & 0.43 & 99.89 & 0.84 & 99.78 & 0.14 & 99.93 & 0.37 & 99.89 \\
  & $M=20$ & EOE & 0.21 & 99.89  & 0.6 & 99.86 & 1.13 & 99.71 & 0.29 & 99.89 & 0.56 & 99.84 \\
  & $M=20$ & Ours & 0.04 & 99.98  & 0.28 & 99.93 & 0.67 & 99.83 & 0.14 & 99.96 & 0.28 & 99.92 \\
        \midrule
        \multirow{4}{*}{\textbf{Average}}
    & $M=10$ & EOE & 5.89 & 98.20 & 0.77 & 99.81 & 1.45 & 99.65 & 0.79 & 99.80 & 2.23 & 99.37 \\
    & \cellcolor{gray!20}\textbf{$M=10$} & \cellcolor{gray!20}\textbf{Ours} & \cellcolor{gray!20}\textbf{5.20} & \cellcolor{gray!20}\textbf{98.41} & \cellcolor{gray!20}\textbf{0.55} & \cellcolor{gray!20}\textbf{99.86} & \cellcolor{gray!20}\textbf{1.10} & \cellcolor{gray!20}\textbf{99.74} & \cellcolor{gray!20}\textbf{0.73} & \cellcolor{gray!20}\textbf{99.82} & \cellcolor{gray!20}\textbf{1.89} & \cellcolor{gray!20}\textbf{99.46}\\
    & $M=20$ & EOE & 5.59 & 98.29 & 0.60 & 99.85 & 1.18 & 99.72 & 0.77 & 99.81 & 2.04 & 99.42\\
    & \cellcolor{gray!20}\textbf{$M=20$} & \cellcolor{gray!20}\textbf{Ours} & \cellcolor{gray!20}\textbf{5.19} & \cellcolor{gray!20}\textbf{98.4} & \cellcolor{gray!20}\textbf{0.44} & \cellcolor{gray!20}\textbf{99.89} & \cellcolor{gray!20}\textbf{0.94} & \cellcolor{gray!20}\textbf{99.78} & \cellcolor{gray!20}\textbf{0.60} & \cellcolor{gray!20}\textbf{99.86} & \cellcolor{gray!20}\textbf{1.79} & \cellcolor{gray!20}\textbf{99.48}\\
		\bottomrule
	\end{tabular}} 
     \label{ood far dataset & N}
 
\end{table*}

\begin{table*}[ht]
	\centering
 \caption{Zero-shot far OOD detection results for food101 as the ID dataset with different broad category number $N$ and $\beta$ when outlier class label number L=6$\times K$.}
    \resizebox{\linewidth}{!}{
	\begin{tabular}{lccccccccccccccccc}
		\toprule
        &  &  & \multicolumn{8}{c}{\textbf{OOD Dataset}} & \multicolumn{2}{c}{\multirow{2}{*}{\textbf{Average}}}  \\
        \multirow{2}{*}{\shortstack{\textbf{Broad Category} \\ \textbf{Number}}}
        & \multirow{2}{*}{$\beta$}
        & \multirow{2}{*}{\textbf{Method}} 
        & \multicolumn{2}{c}{iNaturalist} 
        & \multicolumn{2}{c}{SUN} 
        & \multicolumn{2}{c}{Places} 
        & \multicolumn{2}{c}{Texture} 
        & & &\\
          \cmidrule(r){4-5} \cmidrule(r){6-7} \cmidrule(r){8-9} \cmidrule(r){10-11} \cmidrule(r){12-13}  
		&  & & \multicolumn{1}{c}{\textbf{FPR95}$\downarrow$} 
        & \multicolumn{1}{c}{\textbf{AUROC}$\uparrow$} 
        & \multicolumn{1}{c}{\textbf{FPR95}$\downarrow$} 
        & \multicolumn{1}{c}{\textbf{AUROC}$\uparrow$} 
        & \multicolumn{1}{c}{\textbf{FPR95}$\downarrow$} 
        & \multicolumn{1}{c}{\textbf{AUROC}$\uparrow$} 
        & \multicolumn{1}{c}{\textbf{FPR95}$\downarrow$} 
        & \multicolumn{1}{c}{\textbf{AUROC}$\uparrow$} 
        & \multicolumn{1}{c}{\textbf{FPR95}$\downarrow$} 
        & \multicolumn{1}{c}{\textbf{AUROC}$\uparrow$}\\ 
		\cmidrule(r){1-1} \cmidrule(r){2-2} \cmidrule(r){3-3} 
        \cmidrule(r){4-5} \cmidrule(r){6-7} \cmidrule(r){8-9} \cmidrule(r){10-11} \cmidrule(r){12-13}  
		\multirow{6}{*}{\textbf{N=10}} 
    & 0.25 & EOE & 5.36 & 98.81 & 0.73 & 99.80 & 1.44 & 99.64 & 2.00 & 99.50 & 2.38 & 99.44 \\
  & 0.25 & MM-OOD & 3.56 & 99.30 & 0.54 & 99.87 & 1.25 & 99.71 & 1.90 & 99.53 & 1.81 & 99.60 \\
  & 0.5 & EOE & 7.22 & 98.23 & 1.52 & 99.56 & 2.09 & 99.42 & 2.50 & 99.30 & 3.33 & 99.13 \\
  & 0.5 & MM-OOD & 3.72 & 99.23 & 0.82 & 99.79 & 1.50 & 99.63 & 2.44 & 99.43 & 2.12 & 99.52 \\
  & 0.75 & EOE & 11.88 & 97.11 & 4.49 & 98.89 & 4.84 & 98.78 & 4.77 & 98.68 & 6.50 & 98.37 \\
  & 0.75 & MM-OOD & 4.68 & 98.97  & 1.89 & 99.51 & 2.50 & 99.36 & 3.62 & 99.14 & 3.17 & 99.25 \\
        \midrule
       \multirow{6}{*}{\textbf{N=20}} 
    & 0.25 & EOE & 4.87 & 98.95 & 0.54 & 99.84 & 1.26 & 99.68 & 2.17 & 99.49 & 2.21 & 99.49 \\
  & 0.25 & MM-OOD & 2.27 & 99.54  & 0.32 & 99.91 & 0.94 & 99.77 & 1.61 & 99.60 & 1.29 & 99.70 \\
  & 0.5 & EOE & 6.94 & 98.36 & 1.44 & 99.59 & 2.13 & 99.43 & 3.22 & 99.21 & 3.43 & 99.15 \\
  & 0.5 & MM-OOD & 2.42 & 99.47 & 0.71 & 99.79 & 1.34 & 99.65 & 2.19 & 99.46 & 1.66 & 99.59 \\
  & 0.75 & EOE & 11.73 & 97.17  & 4.99 & 98.86 & 5.57 & 98.69 & 6.72 & 98.38 & 7.25 & 98.28 \\
  & 0.75 & MM-OOD & 3.18 & 99.24 & 2.33 & 99.4 & 2.87 & 99.27 & 3.91 & 99.01 & 3.07 & 99.23 \\
        \midrule
        \multirow{6}{*}{\textbf{N=40}} 
    & 0.25 & EOE & 3.12 & 99.36 & 0.23 & 99.93 & 0.88 & 99.78 & 1.94 & 99.56 & 1.54 & 99.66 \\
  & 0.25 & MM-OOD & 2.85 & 99.43  & 0.27 & 99.92 & 0.89 & 99.78 & 1.55 & 99.64 & 1.39 & 99.69 \\
    & 0.5 & EOE & 4.23 & 99.09 & 0.43 & 99.86 & 1.17 & 99.69 & 2.87 & 99.32 & 2.17 & 99.49 \\
  & 0.5 & MM-OOD & 3.83 & 99.18 & 0.68 & 99.81 & 1.34 & 99.65 & 2.17 & 99.45 & 2.00 & 99.52 \\
  & 0.75 & EOE & 6.69 & 98.47 & 1.20 & 99.64 & 2.15 & 99.40 & 5.66 & 98.67 & 3.93 & 99.04 \\
  & 0.75 & MM-OOD & 6.26 & 98.64  & 2.45 & 99.43 & 3.15 & 99.25 & 4.55 & 98.93 & 4.10 & 99.06 \\
        \midrule
        \multirow{6}{*}{\textbf{Average}}
    & 0.25 & EOE & 4.45 & 99.04 & 0.50 & 99.86 & 1.19 & 99.70 & 2.04 & 99.52 & 2.04 & 99.53 \\
  & 0.25 & MM-OOD & 5.13 & 98.85  & 0.77 & 99.77 & 1.49 & 99.94 & 1.57 & 99.60 & 1.24 & 99.72 \\
  & 0.5 & EOE &  & 98.85  & 0.77 & 99.77 & 1.49  & 99.45 & 4.10 & 98.68 & 3.78 & 99.10 \\
  & 0.5 & MM-OOD & 2.42 & 99.52  & 1.88 & 99.64 & 0.17 & 99.94 & 1.57 & 99.60 & 1.24 & 99.72 \\
  & 0.75 & EOE & 1.88 & 99.64  & 0.21 & 99.94 & 0.79 & 99.80 & 1.55 & 99.63 & 1.11 & 99.75 \\
  & 0.75 & MM-OOD & 7.07 & 98.44  & 1.43 & 99.68 & 2.5  & 99.45 & 4.1 & 98.68 & 3.78 & 99.1 \\
		\bottomrule
	\end{tabular}}
    
     \label{result1}
\end{table*}


\end{document}